\documentclass[10pt,twocolumn,letterpaper]{article}

\usepackage[pagenumbers]{cvpr} 

\usepackage{graphicx}
\usepackage{amsmath}
\usepackage{amssymb}
\usepackage{booktabs}
\usepackage{xcolor,colortbl}
\definecolor{amber}{rgb}{1.0, 0.75, 0.0}

\usepackage[pagebackref,breaklinks,colorlinks]{hyperref}

\usepackage[capitalize]{cleveref}
\crefname{section}{Sec.}{Secs.}
\Crefname{section}{Section}{Sections}
\Crefname{table}{Table}{Tables}
\crefname{table}{Tab.}{Tabs.}

\def\cvprPaperID{****} 
\def\confName{CVPR}
\def\confYear{2022}

\begin{document}

\title{MSTR: Multi-Scale Transformer for End-to-End\\Human-Object Interaction Detection}
\author{Bumsoo Kim\thanks{this work was done in Kakao Brain}\\
LG AI Research\\
{\tt\small bumsoo.kim@lgresearch.ai}
\and
Jonghwan Mun\\
Kakao Brain\\
\and
Kyoung-Woon On\\
Kakao Brain\\
\and
Minchul Shin\\
Kakao Brain\\
\and
Junhyun Lee\\
Korea University\\
\and
Eun-Sol Kim\\
\small Department of Computer Science \\
\small Hanyang University\\
}
\maketitle

\begin{abstract}
Human-Object Interaction (HOI) detection is the task of identifying a set of $\langle$human, object, interaction$\rangle$ triplets from an image.
Recent work proposed transformer encoder-decoder architectures that successfully eliminated the need for many hand-designed components in HOI detection through end-to-end training.
However, they are limited to single-scale feature resolution, providing suboptimal performance in scenes containing humans, objects, and their interactions with vastly different scales and distances.
To tackle this problem, we propose a Multi-Scale TRansformer (MSTR) for HOI detection powered by two novel HOI-aware deformable attention modules called Dual-Entity attention and Entity-conditioned Context attention.
While existing deformable attention comes at a huge cost in HOI detection performance, our proposed attention modules of MSTR learn to effectively attend to sampling points that are essential to identify interactions.
In experiments, we achieve the new state-of-the-art performance on two HOI detection benchmarks.
\end{abstract}

\section{Introduction}
\label{sec:intro}

\begin{figure}
    \centering
    \includegraphics[width=\columnwidth]{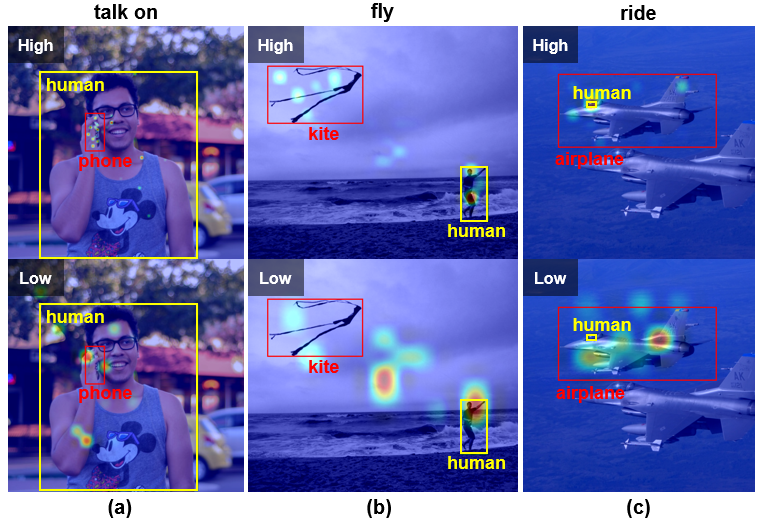}
    \caption{Multi-scale attention of MSTR on interactions including: (a) large human with small object, (b) distant human and object, and (c) small human and a large object.
    The top row (high resolution) and the bottom row (low resolution) captures the context of the interaction in various scales. Best viewed in color.
    }
    \label{fig:attention}
    \vspace{-5pt}
\end{figure}
Human-Object Interaction (HOI) detection is a task to predict a set of $\langle$\textit{human, object, interaction}$\rangle$ triplets in an image~\cite{gupta2015visual}.
Previous methods have indirectly addressed this task by detecting human and object instances and individually inferring interaction labels for every pair of the detected instances with either neural networks (\ie, two-stage HOI detectors~\cite{gkioxari2018detecting,gao2018ican,gupta2019no,qi2018learning,ulutan2020vsgnet,wang2020contextual,liu2020consnet,wang2019deep,peyre2019detecting,xu2019learning,li2020pastanet,gao2020drg,bansal2020detecting,zhong2020polysemy,liu2020amplifying,li2020detailed,li2019transferable,wan2019pose,zhou2019relation}) or triplet matching (\ie, one-stage HOI detectors~\cite{wang2020learning,liao2020ppdm,bkim2020uniondet}).
The additional complexity caused by this indirect inference structure and post-processing (\eg, NMS) stage behaved as a major bottleneck in inference time in HOI detection.
To deal with this bottleneck, transformer-based HOI detectors~\cite{kim2021hotr, tamura2021qpic, chen2021reformulating, zou2021end} have been proposed to achieve end-to-end HOI detection without the need for the post-processing stage mentioned above.
These works have shown competitive performance in both accuracy and inference time with direct set-level prediction and transformer attentions that can exploit the contextual information between humans, objects, and their interactions.

However, due to the huge computational costs raised when processing multi-scale feature maps (with about 20$\times$ more image tokens) with transformer attention, current transformer-based HOI detectors are limited to using only single-scale feature maps.
Due to this limitation, previous transformer-based approaches demonstrate suboptimal performance, especially for scenes where humans, objects, and the contextual information for their interactions exist at various scales.

In this paper, we propose Multi-Scale TRasnformer (MSTR), a transformer-based HOI detector that can exploit multi-scale feature maps for HOI detection.
Inspired by previously proposed deformable attention for standard object detection~\cite{zhu2020deformable}, we aim to efficiently explore multi-scale feature maps by attending to only a small number of sampling points generated from the query element instead of calculating the attention values for the entire spatial dimension.
Yet, we found out in our preliminary experiments that directly applying na\"ive deformable attention in HOI detection leads to a serious performance drop.

To overcome this deterioration, we equipped MSTR with two novel \textit{HOI-aware} deformable attentions, referred by Dual-Entity Attention and Entity-conditioned Context Attention, which are designed to capture the complicated semantics of Human-Object Interaction throughout multi-resolution feature maps (see Figure~\ref{fig:attention}).
Specifically, precise entity-level semantics for humans and objects are captured by \textit{Dual-Entity attention}, while the contextual information for the interaction is conditionally reimbursed by \textit{Entity-conditioned Context attention}.
To further improve performance, we delve into decoder architectures that can effectively handle the multiple semantics obtained from the two HOI-aware attentions above.

The main contributions of our work are threefold:
\begin{itemize}
    \item We propose MSTR, the first transformer-based HOI detector that exploits multi-scale visual feature maps.
    \item We propose new deformable attention modules, called Dual-Entity attention and Entity-conditioned Context attention, which effectively and efficiently capture human, object, and context information associated with HOI queries.
    \item We explore decoder architectures to handle the multiple semantics captured by our proposed deformable attentions and further improve HOI detection performance.
\end{itemize}
\section{Preliminary}
In this section, we start with a basic pipeline of a transformer-based end-to-end HOI detector~\cite{tamura2021qpic}.
Then, we explain the deformable attention module~\cite{zhu2020deformable} that reduces computational cost in attention, thus enabling the transformer to take multi-scale feature maps as an input.
Afterward, we discuss why the direct application of multi-scale deformable attentions is not suitable for HOI detection.
\subsection{End-to-End HOI Detection with Transformers}
Out of the multiple candidates~\cite{kim2021hotr,tamura2021qpic,chen2021reformulating,zou2021end} using transformers for HOI detection, we adopt QPIC~\cite{tamura2021qpic} as our baseline due to its simple structure and good performance.

\paragraph{Set Prediction.}
Transformer-based HOI detectors formulate the task as a set-level prediction problem.
It is achieved by exploiting a fixed number of HOI queries, each of which generates four types of predictions: 1) the coordinate of the human bounding box (\ie, subject of the interaction), 2) the coordinate of the object bounding box (\ie, target of the interaction), 3) the object class and 4) the interaction type.
Note that the set-level predictions are learned using losses based on Hungarian Matching with ground-truths.

\paragraph{Transformer Encoder-Decoder Architecture.}
The architecture of QPIC~\cite{tamura2021qpic} consists of a backbone CNN, a transformer encoder, and a transformer decoder.
Given an image, a single-scale visual feature map is extracted by the backbone CNN, and then positional information is added to the feature map.
The transformer encoder takes the visual features and returns contextualized visual features with self-attention layers.
In the transformer decoder, HOI queries are first processed by the self-attention layer, and then the cross-attention layer associates the HOI queries with the contextualized visual features (given by the encoder) to capture relevant HOI representations.
Finally, predictions for HOI are computed from individual contextualized HOI query embeddings as mentioned above.
Note that both self-attention and cross-attention adopt multi-head attention.

To be specific, given a single-scale input feature map $x\in\mathbb{R}^{C\times H\times W}$ where $C$ is the feature dimension, the single-scale multi-head attention $f^{sg}_q=\mbox{SSAttn}(z_q, x)$ for the $q^{th}$ query feature $z_q$ (either an image token for the encoder or an HOI query for the decoder) is calculated by
\begin{equation}
\label{eq:attn}
    f^{sg}_q=\sum_{m=1}^{M}{W_m}\big[\sum_{k\in\Omega_k}{A_{mqk}\cdot W'_m x_k}\big],
\end{equation}
where $A_{mqk}$ indicates an attention weight calculated with learnable weights $U_m,V_m\in\mathbb{R}^{C_v\times C}$ as $\text{exp}\big(\frac{z_q^TU_m^TV_mx_k}{\sqrt{C_v}}\big)$.
Throughout this paper, for the attention module, we let $m$ index the attention head ($1\leq m \leq M$), $q\in\Omega_q$ indexes a query element with feature $z_q\in\mathbb{R}^C$, $k\in\Omega_k$ indexes a key element with feature $z_k\in\mathbb{R}^C$, while $\Omega_q$ and $\Omega_k$ specify the set of query and key elements, respectively.
$W_m$ and $W'_m$ are learnable embedding parameters for $m^{\text{th}}$ attention head, and $A_{mqk}$ is normalized as $\sum_{k\in\Omega_k}A_{mqk}=1$.

\paragraph{Complexity.}
Given an input feature map $x\in\mathbb{R}^{C\times H\times W}$ and $N$ HOI queries, the complexity of transformer encoder and decoder are $O(H^2W^2C)$ and $O(HWC^2+NHWC+2NC^2+N^2C)$, respectively.
Since the complexity grows in quadratic scale as the spatial resolution ($H$,$W$) increases, it raises significant complexity when exploiting multi-resolution feature maps where there are about 20$\times$ more features to process.

\paragraph{Towards Multi-Scale HOI detection.}
In HOI detection, not only do humans and objects exist at various scales, but they also interact at various distances in images.
Therefore, it is essential to exploit multi-scale feature maps $\{x\}_{l=1}^L$ (where $x^l\in\mathbb{R}^{C\times H_l \times W_l}$, $l$ indexes the feature level) to deal with the various scales of objects and contexts to capture interactions precisely.
However, as multi-scale feature maps have almost $\times 20$ more elements to process than a single-scale feature map, it provokes a serious complexity issue in calculating Eq.~(\ref{eq:attn}).
\subsection{Revisiting Deformable Transformers}
The deformable attention module is proposed to deal with the problem of high complexity in the transformer attention.
The core idea is to reduce the number of \textit{key} elements in the attention module by sampling the small number of spatial locations related to regions of interest for each \textit{query} element.

\paragraph{Sampling Locations for Deformable Attention.}
Given a multi-scale input feature map $\{x^l\}_{l=1}^L$ where $x^l \in \mathbb{R}^{C\times H_l\times W_l}$, the $K$ sampling locations of interest for each attention head and each feature level are generated from each \textit{query} element $z_q\in\mathbb{R}^C$.
Because direct prediction of coordinates of sampling location is difficult to learn, it is formulated as prediction of a reference point $r_q \in [0,1]^{2}$ and $K$ sampling offsets $\Delta r_{q}\in\mathbb{R}^{M\times L\times K \times 2}$.
Then, the $k^{\text{th}}$ sampling location at $l^{\text{th}}$ feature level and $m^{\text{th}}$ attention head for query element $z_q$ is defined by $p_{mlqk}=\phi_{l}(r_q)+\Delta r_{mlqk}$ where $\phi_{l}(\cdot)$ is a function to re-scale the coordinate of reference point to the input feature map of the $l^{\text{th}}$ level.

\paragraph{Deformable Attention Module.}
Given a multi-scale input feature map $\{x^l\}_{l=1}^{L}$, the multi-scale deformable attention  $f^{ms}_q=\mbox{MSDeformAttn}(z_q, p_q, \{x^l\}_{l=1}^{L})$ for query element $z_q$ is calculated using a set of predicted sampling locations $p_q$ as follows:
\begin{equation}
\begin{split}
\label{eq:ms_deform_attn}
    f^{ms}_q=\sum_{m=1}^{M}{W_m}\big[\sum_{l=1}^{L}{\sum_{k=1}^{K}{A_{mlqk}\cdot W'_m\Phi_{mlqk}}}\big],
\end{split}
\end{equation}
where $l$, $k$ and $m$ index the input feature level, the sampling location and the attention head, respectively, while $A_{mlqk}$ indicates an attention weight for the $k^{th}$ sampling location at the $l^{th}$ feature level and the $m^{th}$ attention head.
$\Phi_{mlqk}$ means the sampled $k^{\text{th}}$ key element at $l^{\text{th}}$ feature level and $m^{\text{th}}$ attention head using the sampling location, which is obtained by bilinear interpolation as
$\Phi_{mlqk} = x^{l}(p_{mlqk}) = x^{l}(\phi_{l}(r_q)+\Delta r_{mlqk})$.
Note that for each query element, the attention computation is performed with only sampled regions of interest where the sampled number ($=LMK$) is much smaller than the number of all the key elements ($\sum_{l=1}^L H_lW_l$), thus leads to a reduced computational cost.
\paragraph{Problem with Direct Application to HOI Detection.}
Deformable attention effectively reduces the complexity of exploiting multi-scale features with transformers to an acceptable level.
However, while the sampling procedure above does not deteriorates performance in standard object detection, it causes a serious performance drop in HOI detection (29.07 $\rightarrow$ 25.53) as shown in Table~\ref{tab:ablation}.
We conjecture that this is partly due to the following reasons.
First, unlike the object detection task where an object query is associated with a single object, an HOI query is entangled with multiple semantics (\ie, human, object, and their interaction); thus learning to sample the region of interest for multiple semantics with individual HOI queries (especially with sparse information) is much challenging compared to the counterpart of object detection.
Second, deformable attention is learned to attend only to the sampling points near the localized objects; this leads to the loss of contextual information that is an essential clue for precise HOI detection.
The following sections describe how we resolve these issues and improve performance.
\section{Method}
\label{sec:method}
\begin{figure*}
    \centering
    \includegraphics[width=0.9\textwidth]{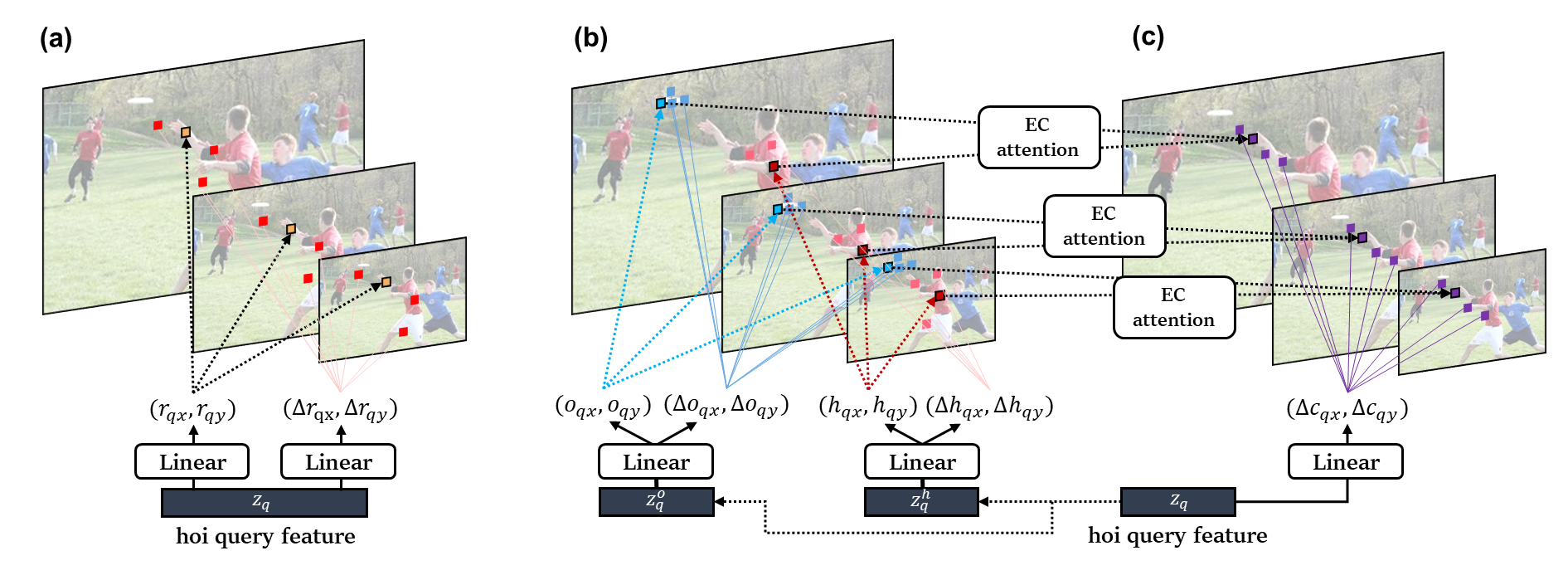}
    \caption{
    Illustration of (a) Deformable Attention, (b) Dual-Entity Attention, (c) Entity-conditioned Context Attention (abbrevieated as EC).
    The sampling point for deformable attention is obtained by combining the reference points with sampling offset.
    In (a), both reference points $r_q=(r_{qx},r_{qy})$ and sampling offsets $\Delta r_{q}=(\Delta r_{qx},\Delta r_{qy})$ are obtained from a single hoi query feature $z_q$.
    In (b), the reference points and sampling offsets for the humans $h_q=(h_{qx},h_{qy})$, $\Delta h_{q}=(\Delta h_{qx},\Delta h_{qy})$ and objects $o_q=(o_{qx},o_{qy})$, $\Delta o_{q}=(\Delta o_{qx},\Delta o_{qy})$ are obtained from $z_q^h$ and $z_q^o$, respectively, which is obtained by a linear projection of $z_q$ (dotted line).
    In (c), the sampling offsets $\Delta c_{q}=(\Delta c_{qx},\Delta c_{qy})$ are obtained from $z_q$ while the reference points are obtained in conditional to entities in (b).
    }
    \label{fig:splitjoint}
\end{figure*}
In this section, we introduce MSTR, a novel deformable transformer architecture that is suitable for multi-scale HOI detection.
To resolve the problems described in our preliminary, MSTR features new \textit{HOI-aware} deformable attentions designed for HOI detection, referred by Dual-Entity attention and Entity-conditioned Context attention.
\subsection{HOI-aware Deformable Attentions}
The objective of our HOI-aware deformable attentions (Dual-Entity attention and Entity-conditioned Context attention) is to efficiently and effectively extract information of HOIs from multi-scale feature maps for a given HOI query.
Figure~\ref{fig:splitjoint} shows conceptual illustrations of (a) deformable attention in literature~\cite{zhu2020deformable}, (b) Dual-Entity attentions and (c) Entity-conditioned Context attention.

\paragraph{Dual-Entity attention for Human\slash Object.}
In HOI detection, the HOI query includes complex and entangled information of multiple semantics: human, object, and interaction information. 
Therefore, it is challenging to accurately predict sampling locations appropriate for each semantic from a single HOI query.
To make sampling locations easier, given an HOI query feature $z_q$, our Dual-Entity attention separately identifies sampling locations for the humans ($p^h_q$) and objects ($p^o_q$).
First, we project $z_q$ with two linear layers to obtain $z_q^h$ and $z_q^o$.
The $k^{\text{th}}$ sampling location at $l^{\text{th}}$ feature level and $m^{\text{th}}$ attention head for human and object are represented by
\begin{equation}
\begin{split}
\label{eq:split_sampling_location}
    p^h_{mlqk}=\phi_{l}(h_q)+\Delta h_{mlqk} 
    ,\\
    p^o_{mlqk}=\phi_{l}(o_q)+\Delta o_{mlqk},
\end{split}
\end{equation}
where $h_q$, $\Delta h$ is the reference point and sampling offsets for humans, and $o_q$, $\Delta o$ is the reference point and sampling offsets for objects, each obtained by a linear projection of $z_q^h$ and $z_q^o$, respectively.
Then, based on the sampled locations, attended features for human ($f_q^h$) and object ($f_q^o$) are computed by
\begin{equation}
\begin{split}
\label{eq:split_attn}
    f_q^h=\text{MSDeformAttn}(z_q^h, p_q^h,\{x^l\}_{l=1}^L),
    \\
    f_q^o=\text{MSDeformAttn}(z_q^o, p_q^o,\{x^l\}_{l=1}^L).
\end{split}
\end{equation}
\paragraph{Entity-conditioned Context attention.}
In HOI detection, contextual information often gives an important clue in identifying interactions.
From this point of view, utilizing the local features obtained from near the human and object regions through the Dual-Entity attention is not sufficient to capture contextual information (see our experimental result in Table \ref{tab:ablation}).
To compensate for this, we define an attention with an additional set of sampling points, namely Entity-conditioned Context attention, that is designed to capture the contextual information in specific.

Given the 2D reference points for the human $h_q=(h_{qx}, h_{qy})$ and the object $o_q=(o_{qx},o_{qy})$, 
the reference point for Entity-conditioned Context attention is conditionally computed with the two references.
Motivated by existing works~\cite{wang2020learning,liao2020ppdm,zhong2021glance}, we define the reference points for interaction as the center of human and object, \ie, $c_q=\big( \frac{h_{qx}+o_{qx}}{2}, \frac{h_{qy}+o_{qy}}{2} \big)$.
Note that we empirically observe that such simple reference points offer competitive performance compared to ones predicted using an additional network, while being much faster.
Then, we predict the sampling offsets $\Delta c_q$ from the HOI query feature, obtaining $p^c_{mlqk}=\phi_l(c_q)+\Delta c_{mlqk}$.
Finally, the attended feature for contextual information $f^c_q$ is computed using sampling location $p_q^c$ as follows:
\begin{equation}
\begin{split}
\label{eq:join_attn}
    f_q^c=\text{MSDeformAttn}(z_q, p_q^c,\{x^l\}_{l=1}^L).
\end{split}
\end{equation}
\begin{figure*}
    \centering
    \includegraphics[width=0.9\textwidth]{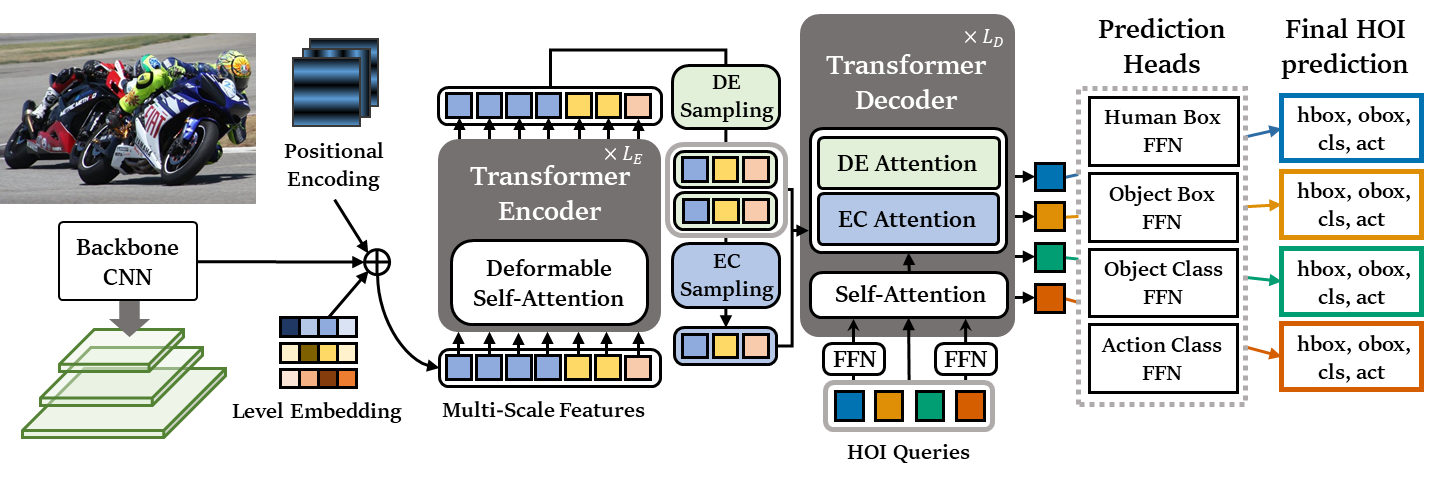}
    \caption{Overall pipeline of MSTR. On top of the standard transformer encoder-decoder architecture for HOI detection (\ie, QPIC), we leverage deformable samplings for the encoder self-attention and the decoder cross-attention modules to deal with the huge complexity caused by using multi-scale feature maps. For the decoder cross-attention, we leverage three sets of key elements sampled for our Dual-Entity attention (denoted as DE sampling, DE attention) and Entity-conditioned Context attention (denoted as EC sampling, EC attention).
    }
    \label{fig:pipeline}
\end{figure*}
\subsection{MSTR Architecture}
\label{sec:architecture}
\begin{figure}
    \centering
    \includegraphics[width=\columnwidth]{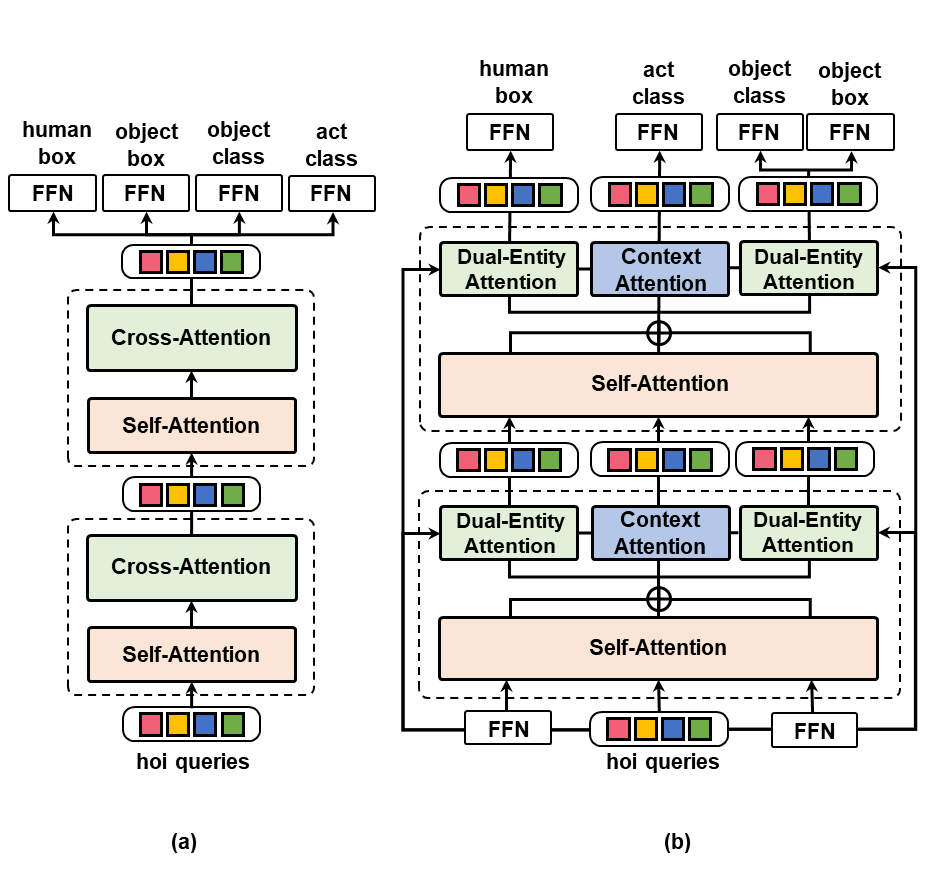}
    \caption{
    Comparison of a simple 2-layer Decoder architecture for Transformer-based HOI detectors: (a) conventional one introduced in QPIC, and (b) HOI-aware one in MSTR.
    Entity-conditioned Context attention is abbreviated as Context Attention.
    MSTR stacks decoder layers by merging the self attention outputs, which further improves performance (see Table~\ref{tab:ablation}).
    }
    \label{fig:architecture}
    \vspace{-15pt}
\end{figure}

In this section, the overall architecture of MSTR with our suggested two deformable attentions will be described (see Figure~\ref{fig:pipeline}).
MSTR follows the previous transformer encoder-decoder architecture, where the encoder performs self-attention given the image features while the decoder performs self-attention for HOI queries followed by cross-attention between updated HOI queries and the encoded image features.

\paragraph{Encoder.}
The encoder of MSTR takes multi-scale input feature maps given by a backbone CNN, performs a series of deformable attention modules in Eq.(\ref{eq:ms_deform_attn}), and finally generates encoded feature maps.
Positional encoding~\cite{carion2020end} is added to preserve spatial information while level embedding~\cite{zhu2020deformable} is added to denote which resolution did the image feature comes from.

\paragraph{Decoder.}
By leveraging our HOI-aware deformable attentions, the cross-attention layer in MSTR decoder extracts three different semantics (human, object, and contextual information) for each HOI query from the encoded image features.
For each decoder layer, we discovered that compositing the multiple semantics obtained from the previous cross-attention layer~\cite{dong2021visual} by summing the semantics after applying individual self-attention demonstrates the best performance (see Table~\ref{tab:ablation} and Appendix).
The input for the $(k+1)$-th layer of our HOI-aware deformable attention $\bar{z}_q^{k+1}$ is written as:
\begin{equation}
\label{eq:split_attn}
    \bar{z}_q^{k+1}=\mbox{SA}(f^h_q(k))+\mbox{SA}(f^o_q(k))+\mbox{SA}(f^c_q(k)),
\end{equation}
where $f^h_q(k)$, $f^o_q(k)$, $f^c_q(k)$ denotes the multiple semantic outputs of the previous ($k$-th) decoder obtained by Eq.(\ref{eq:split_attn}) and Eq.(\ref{eq:join_attn}), respectively. SA denotes Multi-Head Self-Attention operation with Eq.(\ref{eq:attn})~\cite{vaswani2017attention} and $\bar{z}_q^1=\text{SA}(z_q)+\text{SA}(z_q^h)+\text{SA}(z_q^o)$.

\paragraph{MSTR Inference.}
Given the cross attention results of the final decoder layer where $f_q^h$ and $f_q^o$ is obtained by Eq.~(\ref{eq:split_attn}) and $f_q^c$ is obtained by Eq.~(\ref{eq:join_attn}), the final prediction heads in MSTR predict the $\langle \text{bbox}^h_q, \text{bbox}^o_q, \text{cls}^o_q, \text{act}_q \rangle$ using FFN as follows:
\begin{align}
\label{eq:prediction}
    &(u_{qx}, u_{qy},u_{qw},u_{qh})=\mbox{FFN}_{\mbox{hbox}}(f^h_q),
    \\
    &(v_{qx}, v_{qy},v_{qw},v_{qh})=\mbox{FFN}_{\mbox{obox}}(f^o_q),
    \\
    &\mbox{cls}_q = \sigma(\mbox{FFN}_{\mbox{cls}}(f^o_q)),
    \\
    &\mbox{act}_q = \sigma(\mbox{FFN}_{\mbox{act}}(f^c_q)),
\end{align}
where $\text{cls}_q$ and $\text{act}_q$ each denote predictions for object the class and the action class after sigmoid function, and final $\text{bbox}^h_q$ is predicted with the center point $\big(\sigma(u_{qx}+\sigma^{-1}(h_{qx})), \sigma(u_{qy}+\sigma^{-1}(h_{qy}))\big)$, width $u_{qw}$, and height $u_{qh}$.
Likewise, the $\text{bbox}^o_q$ is predicted with center point as $\big(\sigma(v_{qx}+\sigma^{-1}(o_{qx})), \sigma(v_{qy}+\sigma^{-1}(o_{qy}))\big)$, width $v_{qw}$, and height $v_{qh}$.
$\sigma$ and $\sigma^{-1}$ denote the sigmoid and inverse sigmoid function, respectively, and is used to normalize the reference points $h_q,o_q$ and the predicted coordinates of human boxes and object boxes $u_{q\{x,y,w,h\}}, v_{q\{x,y,w,h\}} \in \mathbb{R}$.
\section{Experiment}
\label{sec:exp}
In this section, we show the experimental results of our model in HOI detection.
We first describe the experimental settings such as datasets and evaluation metrics.
Next, we compare MSTR with state-of-the-art works on two different benchmarks (V-COCO and HICO-DET) and provide detailed ablation study for each component.
Through the experiments, we demonstrate that MSTR successfully extends conventional transformer-based HOI detectors to utilize multi-scale feature maps, and emphasize that each component of MSTR contributes to the final HOI detection performance.
Lastly, we provide extensive qualitative results of MSTR.
\subsection{Datasets and Metrics}
We evaluate our model on two widely-used public benchmarks: the V-COCO (\textit{Verbs in COCO})~\cite{gupta2015visual} and HICO-DET~\cite{chao2018learning} datasets.
V-COCO is a subset of COCO composed of 5,400 trainval images and 4,946 test images.
For V-COCO dataset, we report the $\text{AP}_{\text{role}}$ over $25$ interactions in two scenarios.
HICO-DET contains 37,536 and 9,515 images for each training and test split with annotations for 600 $\langle verb, object \rangle$ interaction types.
We follow the previous settings and report the mAP over two evaluation settings (Default and Known Object), each with three different category sets: (1) all 600 HOI categories in HICO (Full), (2) 138 HOI categories with less than 10 training instances (Rare), and (3) 462 HOI categories with 10 or more training instances (Non-Rare).
See Appendix for details of the evaluation settings.
\begin{table}[t!]
\centering
\small
  \begin{tabular}{l|c|c c}
    \toprule
    Method\hspace{70pt} & Backbone & $AP_{\text{role}}^{\#1}$ & $AP_{\text{role}}^{\#2}$ \\ \midrule\hline
    \multicolumn{4}{l}{\textit{Models with \textbf{external features}}} \\ \hline
    TIN (R$\text{P}_{\text{D}}\text{C}_{\text{D}}$)~\cite{li2019transferable} & R50 & 47.8 & - \\
    Verb Embedding~\cite{xu2019learning} & R50 & 45.9 & -\\
    RPNN~\cite{zhou2019relation} & R50 & - & 47.5 \\
    PMFNet~\cite{wan2019pose} & R50-FPN & 52.0 & - \\
    PastaNet~\cite{li2020pastanet} & R50-FPN & 51.0 & 57.5 \\
    PD-Net~\cite{zhong2020polysemy} & R50 & 52.0 & - \\
    ACP~\cite{kim2020detecting} & R152 & 53.0 & - \\
    FCMNet~\cite{liu2020amplifying} & R50 & 53.1 & - \\
    ConsNet~\cite{liu2020consnet} & R50-FPN & 53.2 & - \\
    \hline
    \rowcolor[gray]{0.85}\multicolumn{4}{l}{\textit{\textbf{Sequential HOI Detectors}}} \\ \hline
    VSRL~\cite{gupta2015visual} & R50-FPN & 31.8 & - \\
    InteractNet~\cite{gkioxari2018detecting} & R50-FPN & 40.0 & 48.0 \\
    BAR-CNN~\cite{kolesnikov2019detecting} & R50-FPN & 43.6 & - \\
    GPNN~\cite{qi2018learning} & R152 & 44.0 & - \\
    iCAN~\cite{gao2018ican} & R50 & 45.3 & 52.4 \\
    TIN (R$\text{C}_{\text{D}}$)~\cite{li2019transferable} & R50 & 43.2 & - \\
    DCA~\cite{wang2019deep} & R50 & 47.3 & - \\
    VCL~\cite{hou2020visual} & R50-FPN & 48.3 & - \\
    DRG~\cite{gao2020drg} & R50-FPN & 51.0 & - \\
    VSGNet~\cite{ulutan2020vsgnet} & R152 & 51.8 & 57.0 \\
    IDN~\cite{li2020hoi} & R50 & 53.3 & 60.3 \\ \hline
    \rowcolor[gray]{0.85}\multicolumn{4}{l}{\textit{\textbf{Parallel HOI Detectors}}} \\ \hline
    UnionDet~\cite{bkim2020uniondet} & R50-FPN & 47.5 & 56.2 \\
    IPNet~\cite{wang2020learning} & HG104 & 51.0 & - \\
    HOI Transformer~\cite{zou2021end}$^\dagger$ & R101 & 52.9 & - \\
    ASNet~\cite{chen2021reformulating}$^\dagger$ & R50 & 53.9 & - \\
    GGNet~\cite{zhong2021glance} & HG104 & 54.7 & - \\
    HOTR~\cite{kim2021hotr}$^\dagger$ & R50 & 55.2 & 64.4 \\
    QPIC~\cite{tamura2021qpic}$^\dagger$ & R50 & 58.8 & 61.0 \\ \midrule
    \textbf{\textit{MSTR (Ours)}} & R50 & \textbf{62.0} & \textbf{65.2} \\
    \bottomrule
  \end{tabular}
\caption{Comparison of performance on V-COCO test set. $AP_{\text{role}}^{\#1}$, $AP_{\text{role}}^{\#2}$ denotes the performance under Scenario 1 and Scenario 2 in V-COCO, respectively. $\dagger$ denotes end-to-end HOI detectors with transformers, which are the main baselines for our work.}
\label{tab:V-COCO role}
\end{table}
\begin{table*}[h!]
  \centering
  \small
  \begin{tabular}{l c c c c c c c c c}
    \toprule
    \multicolumn{4}{c}{} & \multicolumn{3}{c}{\textbf{Default}} & \multicolumn{3}{c}{\textbf{Known Object}} \\
    \cmidrule(r){5-7} \cmidrule(r){8-10} 
    Method & Detector & Backbone & Feature & Full & Rare & Non Rare & Full & Rare & Non Rare \\ \midrule\hline
    \rowcolor[gray]{0.85}\multicolumn{10}{l}{\textit{\textbf{Sequential HOI Detectors}}} \\ \hline
    
    \multicolumn{1}{l|}{Functional Gen.~\cite{bansal2020detecting}} & HICO-DET & R101 & \multicolumn{1}{c|}{A+S+L} & 21.96 & 16.43 & 23.62 & - & - & - \\
    \multicolumn{1}{l|}{TIN~\cite{li2019transferable}} & HICO-DET & R50 & \multicolumn{1}{c|}{A+S+P} & 22.90 & 14.97 & 25.26 & - & - & - \\
    \multicolumn{1}{l|}{VCL~\cite{hou2020visual}} & HICO-DET & R50 & \multicolumn{1}{c|}{A+S} & 23.63 & 17.21 & 25.55 & 25.98 & 19.12 & 28.03 \\
    \multicolumn{1}{l|}{ConsNet~\cite{liu2020consnet}} & HICO-DET & R50-FPN & \multicolumn{1}{c|}{A+S+L} & 24.39 & 17.10 & 26.56 & 30.34 & 23.40 & 32.41 \\
    \multicolumn{1}{l|}{DRG~\cite{gao2020drg}} & HICO-DET & R50-FPN & \multicolumn{1}{c|}{A+S} & 24.53 & 19.47 & 26.04 & 27.98 & 23.11 & 29.43 \\
    \multicolumn{1}{l|}{IDN~\cite{li2020hoi}} & HICO-DET & R50 & \multicolumn{1}{c|}{A+S} & 24.58 & 20.33 & 25.86 & 27.89 & 23.64 & 29.16 \\ \midrule\hline
    
    \rowcolor[gray]{0.85}\multicolumn{10}{l}{\textit{\textbf{Parallel HOI Detectors}}} \\ \hline
    \multicolumn{1}{l|}{UnionDet~\cite{bkim2020uniondet}} & HICO-DET & R50-FPN & \multicolumn{1}{c|}{A} & 17.58 & 11.72 & 19.33 & 19.76 & 14.68 & 21.27 \\
    \multicolumn{1}{l|}{PPDM~\cite{liao2020ppdm}} & HICO-DET & HG104 & \multicolumn{1}{c|}{A} & 21.10 & 14.46 & 23.09 & 24.81 & 17.09 & 27.12 \\
    \multicolumn{1}{l|}{HOI Transformer~\cite{zou2021end}$^\dagger$} & HICO-DET & R50 & \multicolumn{1}{c|}{A} & 23.46 & 16.91 & 25.41 & 26.15 & 19.24 & 28.22 \\
    \multicolumn{1}{l|}{HOTR~\cite{kim2021hotr}$^\dagger$} & HICO-DET & R50 & \multicolumn{1}{c|}{A} & 25.10 & 17.34 & 27.42 & - & - & - \\
    \multicolumn{1}{l|}{GGNet~\cite{zhong2021glance}} & HICO-DET & HG104 & \multicolumn{1}{c|}{A} & 28.83 & 22.13 & 30.84 & 27.36 & 20.23 & 29.48 \\
    \multicolumn{1}{l|}{AS-Net~\cite{chen2021reformulating}$^\dagger$} & HICO-DET & R50 & \multicolumn{1}{c|}{A} & 28.87 & 24.25 & 30.25 & 31.74 & 27.07 & 33.14 \\
    \multicolumn{1}{l|}{QPIC~\cite{tamura2021qpic}$^\dagger$} & HICO-DET & R50 & \multicolumn{1}{c|}{A} & 29.07 & 21.85 & 31.23 & 31.68 & 24.14 & 33.93 \\ \midrule
    \multicolumn{1}{l|}{\textit{\textbf{MSTR (Ours)}}} & HICO-DET & R50 &
    \multicolumn{1}{c|}{A} & \textbf{31.17} & \textbf{25.31} & \textbf{32.92} & \textbf{34.02} & \textbf{28.83} & \textbf{35.57} \\
    \bottomrule
  \end{tabular}
  \caption{
  Performance comparison in HICO-DET. The Detector column is denoted as `HICO-DET' to show that the object detector is fine-tuned on the HICO-DET training set. Each letter in Feature column stands for A: Appearance (Visual Features), S: Interaction Patterns (Spatial Correlations), P: Pose Estimation, L: Linguistic Priors, V: Volume. $\dagger$ denotes end-to-end HOI detectors with transformers.
  Note that all the baseline models without $\dagger$ are already based on multi-scale feature maps.
  }
  \label{tab:HICO-DET role}
\end{table*}
\subsection{Quantitative Results}
We use the standard evaluation code~\footnote{https://github.com/YueLiao/PPDM} following the previous works~\cite{kim2021hotr,tamura2021qpic,chen2021reformulating,zou2021end} to calculate metric scores for both V-COCO and HICO-DET.

\paragraph{Comparison to State-of-The-Art.}
We compare MSTR with state-of-the-art methods in Table~\ref{tab:V-COCO role} and Table~\ref{tab:HICO-DET role}.
In Table~\ref{tab:V-COCO role}, MSTR outperforms the previous state-of-the-art method in V-COCO dataset by a large margin (+3.2p in $\text{AP}_{\text{role}}^{\#1}$ and +4.2p in $\text{AP}_{\text{role}}^{\#1}$).
Similar to this, in Table~\ref{tab:HICO-DET role}, MSTR achieves the highest mAP on HICO-DET dataset in all Full, Rare, and Non-Rare classes obtaining +2.1p, +3.46p, and +1.69p gain for each compared to the previous state-of-the-art.
We use the same scoring function as QPIC without any modification for a fair comparison.
Note that MSTR benefits from the advantages of using deformable attention: the fast convergence for training~\cite{zhu2020deformable} (see more details and the convergence graph in our Appendix).
\begin{table}[t!]
  \centering
  \small
  \scalebox{0.95}{
  \begin{tabular}{l | c c c c c c }
    \toprule
    Method & MS & DA & DE & EC & mAP \\ \hline
    (a) QPIC & & & & & 29.07 \\
    (b) SS-Baseline & & \checkmark & & & 25.53 \\
    (c) SS-Baseline + DE & & \checkmark & \checkmark & & 27.06 \\
    (d) SS-Baseline + DE + EC & & \checkmark & \checkmark & \checkmark & 27.70 \\ \hline
    (e) MS-Baseline & \checkmark & \checkmark & & & 27.52 \\
    (f) MS-Baseline + DE & \checkmark & \checkmark & \checkmark & & 28.30 \\
    (g) MS-Baseline + DE + EC & \checkmark & \checkmark & \checkmark & \checkmark & 30.14 \\
    \textit{\textbf{(h) MSTR (Ours)}} & \checkmark & \checkmark & \checkmark & \checkmark & \textbf{31.17} \\
    \bottomrule
  \end{tabular}
  }
  \caption{
  Comparison of MSTR with our baseline QPIC and its variants in the HICO-DET test set.
  SS and MS denote the models using single scale feature map and multi-scale feature maps, respectively.
  DE and EC indicate our proposed Dual-Entity attention and Entity-conditioned Context attention, respectively.
  }
  \label{tab:ablation}\
  \vspace{-15pt}
\end{table}
\subsection{Ablation Study}
We perform ablations to check the effects of our proposed Dual-Entity attention, Entity-conditioned Context attention, and our proposed decoder architecture that merges the self-attention of the multiple semantics.

\paragraph{Baselines.}
On basis of QPIC~\cite{tamura2021qpic} structure, we define several variants for baselines by applying different combinations of sub-components from MSTR: \textit{multi-scale feature maps (MS)}, \textit{Deformable Attention (DA)}, \textit{Dual-Entity attention (DE)}, and \textit{Entity-conditioned Context attention (EC)}.
Specifically, since deformable attention can be also applied to a single-scale feature map, \textit{SS-Baseline} denotes QPIC where the attention in the transformer is replaced by DA.
Our work can be seen as a process of improving the score to the state-of-the-art by adapting \textit{MS, DE, EC} step by step to \textit{SS-Baseline}.
\textit{MS-Baseline+DE+EC} represents MSTR without merging with self-attention, instead simply passing the sum of the outputs to the next decoder layer.

\paragraph{HOI-Aware Deformable Attentions.}
In Table~\ref{tab:ablation}, we explore the effect of our proposed HOI-Aware Deformable Attentions: Dual-Entity attention and Entity-conditioned Context attention.
As deformable attentions can also be applied in a single-scale feature map, we verify the effectiveness of our proposed deformable attentions on both single-scale and multi-scale baselines.
As we described in our preliminary, the na\"ive implementation of deformable attention on top of QPIC (for single-scale) significantly degrades the score in both single-scale and multi-scale environments (see (a vs. b) and (a vs. e)).
The use of Dual-Entity attention (DE) consistently improves the score in both single-scale (+1.53p in (b vs. c)) and multi-scale environments (+0.78p in (e vs. f)).
As well, Entity-conditioned Context attention (EC) contributes in the multi-scale environment when jointly used with DE (+0.64p in SS and +1.84p in MS).
Therefore, we conclude that disentangling the references (DE) and conditionally reimbursing context information (EC) each gradually contributes to the final performance of HOI detection in both single-scale and multi-scale environments, enabling MSTR to effectively explore multi-scale feature maps to achieve state-of-the-art performance.

\paragraph{Single-scale vs. Multi-scale.}
In Table~\ref{tab:V-COCO role} and Table~\ref{tab:HICO-DET role}, we demonstrate that our method using the multi-scale feature maps outperform all previous methods, including transformer-based methods~\cite{kim2021hotr,chen2021reformulating,tamura2021qpic,zou2021end} and the ones that already use multi-scale feature maps heavily~\cite{bkim2020uniondet,gao2020drg,hou2020visual,zhong2021glance,li2020hoi,liu2020consnet}.
To analyze further, Table~\ref{tab:ablation} compares single-scale version and the multi-scale version of our baselines (see (b-e) and (e-h)).
In all cases of converting the single-scale feature map to the multi-scale one, we observe consistent performance gains (see (b vs. e), (c vs. f), and (d vs. g,h)).
The gain is maximized when \textit{DE} and \textit{EC} are used together.
We further provide a detailed analysis of the effectiveness of MSTR in multi-scale environments in our Appendix.

\paragraph{Decoder Architecture.}
We verify the effectiveness of Figure~\ref{fig:architecture} (b) architecture in Table~\ref{tab:ablation} (g vs. h). 
As MSTR considers multiple semantics with two suggested deformable modules, it is important to find suitable decoder architecture which can effectively merge the semantics~\cite{dong2021visual}.
According to the possible combination ways when merging three kinds of semantics, various types of decoder architecture can be candidates for the decoder architectures (described in Appendix).
In our Appendix, we empirically verify that Figure~\ref{fig:architecture} (b) architecture shows the most powerful and robust performance across all datasets.
\begin{figure}
    \centering
    \includegraphics[width=0.9\columnwidth]{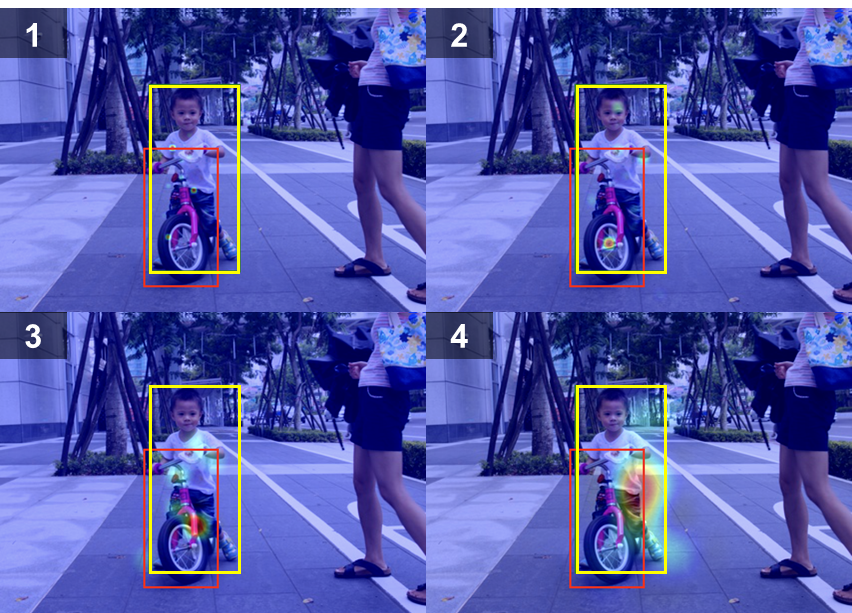}
    \caption{
    Visualization of our Entity-conditioned Context attentions on different levels of feature map (1 being the highest and 4 being the lowest resolution).
    Best viewed in color and scale.
    }
    \label{fig:feature_scale}
\end{figure}
\begin{figure}
    \centering
    \includegraphics[width=0.9\columnwidth]{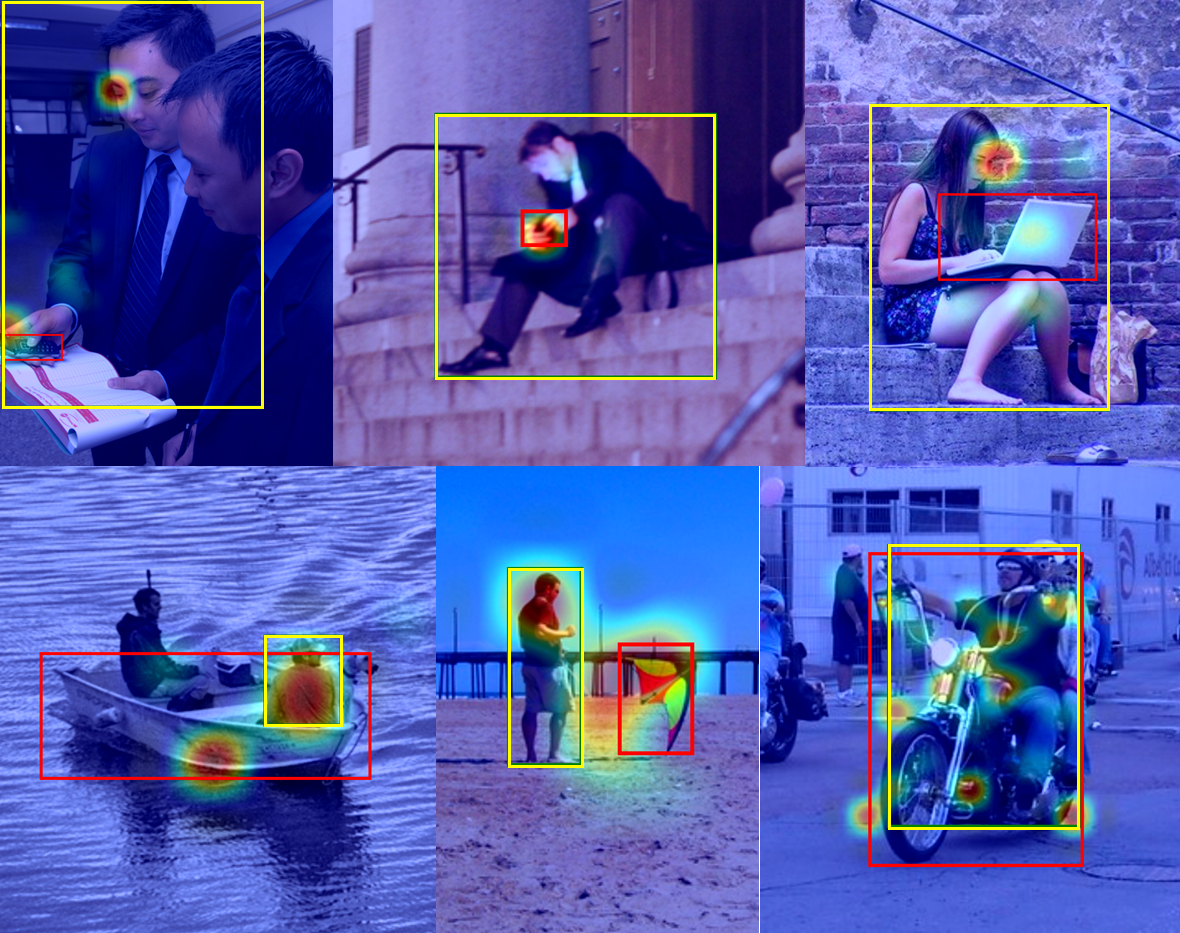}
    \caption{Visualization of the HOI-aware attention of MSTR on different scales of humans and objects.}
    \label{fig:object_scale}
\end{figure}
\subsection{Qualitative Results}
We conduct qualitative analysis of MSTR to observe how MSTR captures interactions.
Figure~\ref{fig:attention} and
Figure~\ref{fig:feature_scale} show the visualization of the attention map in MSTR in various feature levels.
Interestingly, we can observe that in the higher resolution feature maps, the sampling points capture the detail of the interacting human and object while the lower resolution feature maps tend to capture the overall pose or context of the interaction.
In Figure~\ref{fig:attention} and Figure~\ref{fig:object_scale}, we can observe how MSTR attends to test images that include various scales of humans, target objects, and distances.
More details along with quantitative results will be provided in our Appendix.
\section{Related Work}
\label{sec:related_work}
\paragraph{Transformer Based HOI Detectors.}
Human-Object Interaction detection has been initially proposed in \cite{gupta2015visual}, and has been developed in two main streams: sequential methods~\cite{gkioxari2018detecting,gao2018ican,gupta2019no,qi2018learning,ulutan2020vsgnet,wang2020contextual,liu2020consnet,wang2019deep,peyre2019detecting,xu2019learning,li2020pastanet,gao2020drg,bansal2020detecting,zhong2020polysemy,liu2020amplifying,li2020detailed,li2019transferable,wan2019pose,zhou2019relation} and parallel methods~\cite{wang2020learning,liao2020ppdm,bkim2020uniondet}.
However, since these works required hand-crafted post-processing, HOI detectors with transformers have been proposed to eliminate the post-processing step through an end-to-end fashioned set prediction approach~\cite{kim2021hotr,tamura2021qpic,chen2021reformulating,zou2021end}.
Yet, all these methods are limited to a single-scale feature map due to the complexity caused when processing multi-scale feature maps with transformer attention.
\vspace{-0.15cm}

\paragraph{Deformable Transformers for Object Detection.}
DETR has been recently proposed to eliminate the need for many hand-designed components in object detection~\cite{carion2020end}.
Deformable DETR~\cite{zhu2020deformable} mitigates the slow convergence and high complexity issues of DETR and successfully exploits multi-resolution feature maps.
The deformable attention modules in~\cite{zhu2020deformable} attend to a small set of sampling locations as a pre-filter for prominent key elements out of all the feature map pixels.
However, unlike object detection, we observed that this pre-filter seriously deteriorates performance when applied to HOI detection.
Therefore, in this paper, we focus on finding a proper way to incorporate deformable attention into HOI detection for exploiting multi-scale feature maps.
\section{Conclusion}
\label{sec:conc}
In this paper, we present MSTR, the first multi-scale approach in transformer-based HOI detectors.
MSTR overcomes the issues of extending transfomer-based HOI detectors to multi-scale feature maps with novel HOI-Aware Deformable attentions named as Dual-Entity attention and Entity-conditioned Context attention.
In virtue of the two attention modules and our decoder architecture that effectively collects the multiple semantics from each of the attentions, MSTR achieves the state-of-the-art performance in two benchmark datasets in HOI detection.

{\small
\bibliographystyle{ieee_fullname}
\bibliography{cvpr}

\begin{thebibliography}{10}\itemsep=-1pt

\bibitem{bansal2020detecting}
Ankan Bansal, Sai~Saketh Rambhatla, Abhinav Shrivastava, and Rama Chellappa.
\newblock Detecting human-object interactions via functional generalization.
\newblock In {\em AAAI}, pages 10460--10469, 2020.

\bibitem{carion2020end}
Nicolas Carion, Francisco Massa, Gabriel Synnaeve, Nicolas Usunier, Alexander
  Kirillov, and Sergey Zagoruyko.
\newblock End-to-end object detection with transformers.
\newblock {\em arXiv preprint arXiv:2005.12872}, 2020.

\bibitem{chao2018learning}
Yu-Wei Chao, Yunfan Liu, Xieyang Liu, Huayi Zeng, and Jia Deng.
\newblock Learning to detect human-object interactions.
\newblock In {\em 2018 ieee winter conference on applications of computer
  vision (wacv)}, pages 381--389. IEEE, 2018.

\bibitem{chen2021reformulating}
Mingfei Chen, Yue Liao, Si Liu, Zhiyuan Chen, Fei Wang, and Chen Qian.
\newblock Reformulating hoi detection as adaptive set prediction.
\newblock In {\em Proceedings of the IEEE/CVF Conference on Computer Vision and
  Pattern Recognition}, pages 9004--9013, 2021.

\bibitem{dong2021visual}
Qi Dong, Zhuowen Tu, Haofu Liao, Yuting Zhang, Vijay Mahadevan, and Stefano
  Soatto.
\newblock Visual relationship detection using part-and-sum transformers with
  composite queries.
\newblock In {\em Proceedings of the IEEE/CVF International Conference on
  Computer Vision}, pages 3550--3559, 2021.

\bibitem{gao2020drg}
Chen Gao, Jiarui Xu, Yuliang Zou, and Jia-Bin Huang.
\newblock Drg: Dual relation graph for human-object interaction detection.
\newblock In {\em European Conference on Computer Vision}, pages 696--712.
  Springer, 2020.

\bibitem{gao2018ican}
Chen Gao, Yuliang Zou, and Jia-Bin Huang.
\newblock ican: Instance-centric attention network for human-object interaction
  detection.
\newblock {\em arXiv preprint arXiv:1808.10437}, 2018.

\bibitem{gkioxari2018detecting}
Georgia Gkioxari, Ross Girshick, Piotr Doll{\'a}r, and Kaiming He.
\newblock Detecting and recognizing human-object interactions.
\newblock In {\em Proceedings of the IEEE Conference on Computer Vision and
  Pattern Recognition}, pages 8359--8367, 2018.

\bibitem{gupta2015visual}
Jitendra Gupta, Saurabh~Malik.
\newblock Visual semantic role labeling.
\newblock {\em arXiv preprint arXiv:1505.04474}, 2015.

\bibitem{gupta2019no}
Tanmay Gupta, Alexander Schwing, and Derek Hoiem.
\newblock No-frills human-object interaction detection: Factorization, layout
  encodings, and training techniques.
\newblock In {\em Proceedings of the IEEE International Conference on Computer
  Vision}, pages 9677--9685, 2019.

\bibitem{he2016deep}
Kaiming He, Xiangyu Zhang, Shaoqing Ren, and Jian Sun.
\newblock Deep residual learning for image recognition.
\newblock In {\em Proceedings of the IEEE conference on computer vision and
  pattern recognition}, pages 770--778, 2016.

\bibitem{hou2020visual}
Zhi Hou, Xiaojiang Peng, Yu Qiao, and Dacheng Tao.
\newblock Visual compositional learning for human-object interaction detection.
\newblock {\em arXiv preprint arXiv:2007.12407}, 2020.

\bibitem{bkim2020uniondet}
Bumsoo Kim, Taeho Choi, Jaewoo Kang, and Hyunwoo Kim.
\newblock Uniondet: Union-level detection towards real-time human-object
  interaction detection.
\newblock In {\em Proceedings of the European conference on computer vision
  (ECCV)}, 2020.

\bibitem{kim2021hotr}
Bumsoo Kim, Junhyun Lee, Jaewoo Kang, Eun-Sol Kim, and Hyunwoo~J Kim.
\newblock Hotr: End-to-end human-object interaction detection with
  transformers.
\newblock In {\em Proceedings of the IEEE/CVF Conference on Computer Vision and
  Pattern Recognition}, pages 74--83, 2021.

\bibitem{kim2020detecting}
Dong-Jin Kim, Xiao Sun, Jinsoo Choi, Stephen Lin, and In~So Kweon.
\newblock Detecting human-object interactions with action co-occurrence priors.
\newblock {\em arXiv preprint arXiv:2007.08728}, 2020.

\bibitem{kolesnikov2019detecting}
Alexander Kolesnikov, Alina Kuznetsova, Christoph Lampert, and Vittorio
  Ferrari.
\newblock Detecting visual relationships using box attention.
\newblock In {\em Proceedings of the IEEE International Conference on Computer
  Vision Workshops}, pages 0--0, 2019.

\bibitem{li2020detailed}
Yong-Lu Li, Xinpeng Liu, Han Lu, Shiyi Wang, Junqi Liu, Jiefeng Li, and Cewu
  Lu.
\newblock Detailed 2d-3d joint representation for human-object interaction.
\newblock In {\em Proceedings of the IEEE/CVF Conference on Computer Vision and
  Pattern Recognition}, pages 10166--10175, 2020.

\bibitem{li2020hoi}
Yong-Lu Li, Xinpeng Liu, Xiaoqian Wu, Yizhuo Li, and Cewu Lu.
\newblock Hoi analysis: Integrating and decomposing human-object interaction.
\newblock {\em Advances in Neural Information Processing Systems}, 33, 2020.

\bibitem{li2020pastanet}
Yong-Lu Li, Liang Xu, Xinpeng Liu, Xijie Huang, Yue Xu, Shiyi Wang, Hao-Shu
  Fang, Ze Ma, Mingyang Chen, and Cewu Lu.
\newblock Pastanet: Toward human activity knowledge engine.
\newblock In {\em Proceedings of the IEEE/CVF Conference on Computer Vision and
  Pattern Recognition}, pages 382--391, 2020.

\bibitem{li2019transferable}
Yong-Lu Li, Siyuan Zhou, Xijie Huang, Liang Xu, Ze Ma, Hao-Shu Fang, Yanfeng
  Wang, and Cewu Lu.
\newblock Transferable interactiveness knowledge for human-object interaction
  detection.
\newblock In {\em Proceedings of the IEEE Conference on Computer Vision and
  Pattern Recognition}, pages 3585--3594, 2019.

\bibitem{liao2020ppdm}
Yue Liao, Si Liu, Fei Wang, Yanjie Chen, Chen Qian, and Jiashi Feng.
\newblock Ppdm: Parallel point detection and matching for real-time
  human-object interaction detection.
\newblock In {\em Proceedings of the IEEE/CVF Conference on Computer Vision and
  Pattern Recognition}, pages 482--490, 2020.

\bibitem{liu2020amplifying}
Y Liu, Q Chen, and A Zisserman.
\newblock Amplifying key cues for human-object-interaction detection.
\newblock {\em Lecture Notes in Computer Science}, 2020.

\bibitem{liu2020consnet}
Ye Liu, Junsong Yuan, and Chang~Wen Chen.
\newblock Consnet: Learning consistency graph for zero-shot human-object
  interaction detection.
\newblock In {\em Proceedings of the 28th ACM International Conference on
  Multimedia}, pages 4235--4243, 2020.

\bibitem{peyre2019detecting}
Julia Peyre, Ivan Laptev, Cordelia Schmid, and Josef Sivic.
\newblock Detecting unseen visual relations using analogies.
\newblock In {\em Proceedings of the IEEE International Conference on Computer
  Vision}, pages 1981--1990, 2019.

\bibitem{qi2018learning}
Siyuan Qi, Wenguan Wang, Baoxiong Jia, Jianbing Shen, and Song-Chun Zhu.
\newblock Learning human-object interactions by graph parsing neural networks.
\newblock In {\em Proceedings of the European Conference on Computer Vision
  (ECCV)}, pages 401--417, 2018.

\bibitem{tamura2021qpic}
Masato Tamura, Hiroki Ohashi, and Tomoaki Yoshinaga.
\newblock Qpic: Query-based pairwise human-object interaction detection with
  image-wide contextual information.
\newblock In {\em Proceedings of the IEEE/CVF Conference on Computer Vision and
  Pattern Recognition}, pages 10410--10419, 2021.

\bibitem{ulutan2020vsgnet}
Oytun Ulutan, ASM Iftekhar, and Bangalore~S Manjunath.
\newblock Vsgnet: Spatial attention network for detecting human object
  interactions using graph convolutions.
\newblock In {\em Proceedings of the IEEE/CVF Conference on Computer Vision and
  Pattern Recognition}, pages 13617--13626, 2020.

\bibitem{vaswani2017attention}
Ashish Vaswani, Noam Shazeer, Niki Parmar, Jakob Uszkoreit, Llion Jones,
  Aidan~N Gomez, {\L}ukasz Kaiser, and Illia Polosukhin.
\newblock Attention is all you need.
\newblock In {\em Advances in neural information processing systems}, pages
  5998--6008, 2017.

\bibitem{wan2019pose}
Bo Wan, Desen Zhou, Yongfei Liu, Rongjie Li, and Xuming He.
\newblock Pose-aware multi-level feature network for human object interaction
  detection.
\newblock In {\em Proceedings of the IEEE International Conference on Computer
  Vision}, pages 9469--9478, 2019.

\bibitem{wang2020contextual}
Hai Wang, Wei-shi Zheng, and Ling Yingbiao.
\newblock Contextual heterogeneous graph network for human-object interaction
  detection.
\newblock {\em arXiv preprint arXiv:2010.10001}, 2020.

\bibitem{wang2019deep}
Tiancai Wang, Rao~Muhammad Anwer, Muhammad~Haris Khan, Fahad~Shahbaz Khan,
  Yanwei Pang, Ling Shao, and Jorma Laaksonen.
\newblock Deep contextual attention for human-object interaction detection.
\newblock {\em arXiv preprint arXiv:1910.07721}, 2019.

\bibitem{wang2020learning}
Tiancai Wang, Tong Yang, Martin Danelljan, Fahad~Shahbaz Khan, Xiangyu Zhang,
  and Jian Sun.
\newblock Learning human-object interaction detection using interaction points.
\newblock In {\em Proceedings of the IEEE/CVF Conference on Computer Vision and
  Pattern Recognition}, pages 4116--4125, 2020.

\bibitem{wang2021pnp}
Tao Wang, Li Yuan, Yunpeng Chen, Jiashi Feng, and Shuicheng Yan.
\newblock Pnp-detr: Towards efficient visual analysis with transformers.
\newblock In {\em Proceedings of the IEEE/CVF International Conference on
  Computer Vision}, pages 4661--4670, 2021.

\bibitem{xu2019learning}
Bingjie Xu, Yongkang Wong, Junnan Li, Qi Zhao, and Mohan~S Kankanhalli.
\newblock Learning to detect human-object interactions with knowledge.
\newblock In {\em Proceedings of the IEEE Conference on Computer Vision and
  Pattern Recognition}, 2019.

\bibitem{zhong2020polysemy}
Xubin Zhong, Changxing Ding, Xian Qu, and Dacheng Tao.
\newblock Polysemy deciphering network for human-object interaction detection.
\newblock In {\em Proc. Eur. Conf. Comput. Vis}, 2020.

\bibitem{zhong2021glance}
Xubin Zhong, Xian Qu, Changxing Ding, and Dacheng Tao.
\newblock Glance and gaze: Inferring action-aware points for one-stage
  human-object interaction detection.
\newblock In {\em Proceedings of the IEEE/CVF Conference on Computer Vision and
  Pattern Recognition}, pages 13234--13243, 2021.

\bibitem{zhou2019relation}
Penghao Zhou and Mingmin Chi.
\newblock Relation parsing neural network for human-object interaction
  detection.
\newblock In {\em Proceedings of the IEEE International Conference on Computer
  Vision}, pages 843--851, 2019.

\bibitem{zhu2020deformable}
Xizhou Zhu, Weijie Su, Lewei Lu, Bin Li, Xiaogang Wang, and Jifeng Dai.
\newblock Deformable detr: Deformable transformers for end-to-end object
  detection.
\newblock {\em arXiv preprint arXiv:2010.04159}, 2020.

\bibitem{zou2021end}
Cheng Zou, Bohan Wang, Yue Hu, Junqi Liu, Qian Wu, Yu Zhao, Boxun Li, Chenguang
  Zhang, Chi Zhang, Yichen Wei, et~al.
\newblock End-to-end human object interaction detection with hoi transformer.
\newblock In {\em Proceedings of the IEEE/CVF Conference on Computer Vision and
  Pattern Recognition}, pages 11825--11834, 2021.

\end{thebibliography}
}

\appendix
\clearpage
\section{Appendix}
In this Appendix, we provide
\textbf{i)} extended quantitative analysis of MSTR capturing HOI detection in a multi-scale environment,
\textbf{ii)} exploration for various possible decoder architectures,
\textbf{iii)} implementation details of MSTR,
\textbf{iv)} details on experimental datasets and metrics, 
\textbf{v)} details of training,
\textbf{vi)} analysis on convergence speed,
\textbf{vii)} additional qualitative result on our Dual-Entity attention and Entity-conditioned Context attention,
and finally, \textbf{viii)} limitations of our work.
\subsection{Additional Quantitative Results for MSTR}
\label{sec:multi}

First, we perform an extended quantitative analysis on the HICO-DET test set to validate the effectiveness of MSTR in a multi-scale environment.
MSTR uses multi-scale feature maps to explore the semantics of HOI existing in different scales.
In this section, we provide extensive quantitative results that shows the effectiveness of MSTR in capturing the interactions between humans and objects not only at different scales, but in various distances also (\eg, \textit{adjacent} interaction such as `holding a book' or \textit{remote} interaction such as `throwing a frisbee'). 
To this end, we show quantitative results for multi-scale interactions according to 1) relative area of the human and the object, 2) the size of humans/objects, 3) distance between the human and object.
For each criterion, we measure the performance across three bins where each bin has an equal and sufficient amount of HOI ground-truth labels to cover ($\sim$11,000 HOIs).
For comparison, we set QPIC~\cite{tamura2021qpic}, the state-of-the-art transformer-based approach that uses a single-scale feature map, as our baseline.
Note that in this appendix, the size, area, and distance are all calculated in \textit{normalized} image coordinates.

\paragraph{Relative area of human vs.~object.}
To observe how MSTR handles interaction between humans and objects with different scales, we first calculate the average precision (AP) over interaction labels that have different relative areas of humans and objects ($\frac{\text{area(hbox)}}{\text{area(obox)}}$).
We cover three main cases according to their relative areas: i) $\text{AP}_{h<o}$ where the object area is significantly larger than the human area (\eg, human \textit{sitting} on a \textit{bench}), ii) $\text{AP}_{h=o}$ where the human and the object exists in comparable sizes, and iii) $\text{AP}_{h>o}$ where the object area is significantly smaller than the human area (\eg, human \textit{throwing} a \textit{ball}).
We set the threshold for the relative areas so that each bin has an equal number of ground-truth instances (\ie, $\frac{\text{area(hbox)}}{\text{area(obox)}}<0.48$ for $\text{AP}_{h<o}$ and $\frac{\text{area(hbox)}}{\text{area(obox)}}>4.33$ for $\text{AP}_{h>o}$).
In Table~\ref{tab:scale}, MSTR outperforms QPIC in all three types of interaction categories.
Note that the improvement is more substantial in cases where the human and object have vastly different scales (+3.01p for $\text{AP}_{h<o}$ and +1.85p for $\text{AP}_{h>o}$), verifying that MSTR is effectively utilizing multi-scale feature maps.
\begin{table}[t!]
  \centering
  \small
  \begin{tabular}{l c c c}
    \toprule
    Method\hspace{15pt} & $\text{AP}_{h<o}$ & $\text{AP}_{h=o}$ & $\text{AP}_{h>o}$ \\ \midrule
    QPIC & 34.10 & 30.57 & 25.22 \\
    \textbf{MSTR} & \textbf{37.11} & \textbf{31.68} & \textbf{27.07} \\
    \rowcolor[gray]{0.85}$\Delta$AP & \textbf{+3.01} & +1.11 & \textbf{+1.85} \\
    \bottomrule
  \end{tabular}
  \caption{
  Comparison of MSTR with QPIC under interactions with different human/object scale ratio.
  }
  \label{tab:scale}
\end{table}

\paragraph{Human \& object size.}
Here, we compare the average precision over the sizes of humans and objects.
$\text{AP}_\text{L}$, $\text{AP}_\text{M}$, $\text{AP}_\text{S}$ each denotes the average precision for \textbf{L}arge, \textbf{M}iddle, and \textbf{S}mall humans and objects.
In Table~\ref{tab:ho}, MSTR outperforms QPIC in all three categories in both human and object scales.
For the human scales, the improvement is more recognizable in interactions including small human areas (+3.06p in $\text{AP}_\text{S}$) while for object scales, the improvement is consistent over all three scales.
\begin{table}[h!]
  \centering
  \small
  \begin{tabular}{l c c c c c c}
    \toprule
    \multicolumn{1}{c}{} & \multicolumn{3}{c}{\textbf{Human Size}} & \multicolumn{3}{c}{\textbf{Object Size}} \\
    \cmidrule(r){2-4} \cmidrule(r){5-7} 
    Method & $\text{AP}_\text{L}$ & $\text{AP}_\text{M}$ & $\text{AP}_\text{S}$ & $\text{AP}_\text{L}$ & $\text{AP}_\text{M}$ & $\text{AP}_\text{S}$ \\ \midrule
    QPIC & 28.65 & 35.36 & 24.14 & 33.09 & 28.65 & 24.87 \\
    \textbf{MSTR} & \textbf{30.04} & \textbf{37.02} & \textbf{27.20} & \textbf{34.87} & \textbf{30.48} & \textbf{26.60} \\
    \rowcolor[gray]{0.85}$\Delta$AP & +1.39 & +1.66 & \textbf{+3.06} & +1.78 & +1.83 & +1.73 \\
\bottomrule
\end{tabular}
\caption{
Comparison of MSTR with QPIC under different sizes of humans and objects.
}
\label{tab:ho}
\end{table}

\paragraph{Interactions in various distances.}
Not only does MSTR capture interactions with various sized participants, but MSTR also captures interactions with various sized contexts, \ie, interaction in various distances.
To correctly measure how \textit{remote} an interaction is, we note that the distance between center points~\cite{tamura2021qpic} should be normalized by both the image size and the size of the human and object box participating in the interaction.
Given the interaction between hbox ($hx_1,hy_1,hx_2,hy_2$) and obox ($ox_1,oy_1,ox_2,oy_2$), the normalized box area as area (hbox) and area (obox), we define the distance $d_{\text{interaction}}$ as
\begin{align}
\begin{split}
\label{eq:distance}
    &d_{\footnotesize\mbox{center}} = \scriptstyle\sqrt{\left(\frac{hx_1+hx_2}{2}-\frac{ox_1+ox_2}{2}\right)^2 + \left(\frac{hy_1+hy_2}{2}-\frac{oy_1+oy_2}{2}\right)^2}, \\
    &d_{\footnotesize\mbox{interaction}} = d_{\footnotesize\mbox{center}} / (\footnotesize\mbox{area(hbox)}\cdot\mbox{area(obox)}).
\end{split}
\end{align}
Then, we measure the average precision over three categories: i) $\text{AP}_{\text{adjacent}}$ where the human is interacting with a nearby object, ii) $\text{AP}_{\text{distant}}$ where the interacting human/object is within moderate distance, and $\text{AP}_{\text{remote}}$ where the human is interacting with an object sufficiently far away.
As in previous sections, we set the distance threshold so that each bin has an equal number of ground-truth instances.
Table~\ref{tab:distance} shows the improvement of MSTR over QPIC.
Note that while MSTR shows improvement across all three categories, the improvement is more distinguishable in cases where humans are interacting with objects in considerable distance (+2.23p for $\text{AP}_{\text{distant}}$ and +1.89p for $\text{AP}_{\text{remote}}$, respectively).
\begin{table}[t!]
  \centering
  \small
  \begin{tabular}{l c c c}
    \toprule
    Method\hspace{15pt} & $\text{AP}_\text{adjacent}$ & $\text{AP}_\text{distant}$ &
    $\text{AP}_\text{remote}$ \\ \midrule
    QPIC & 31.09 & 31.25 & 21.81 \\
    \textbf{MSTR} & \textbf{32.66} & \textbf{33.48} & \textbf{23.70} \\
    \rowcolor[gray]{0.85}$\Delta$AP & +1.57 & \textbf{+2.23} & \textbf{+1.89} \\
    \bottomrule
  \end{tabular}
  \caption{
  Comparison of MSTR with QPIC under interactions with various distances.
  }
  \label{tab:distance}
\end{table}
\subsection{Analysis on Decoder Architecture}
\label{sec:decod}
As MSTR considers multiple semantics with two suggested deformable modules (Dual-Entity attention and Entity-conditioned Context attention), it is important to find a suitable decoder architecture that effectively merges the semantics.
Here, we explore the possible combinations and various types of decoder architecture candidates when merging the three kinds of semantics.
We empirically verify that MSTR architecture shows the most powerful performance.

\paragraph{Architecture for Dual-Entity attention.}
\begin{figure}
    \centering
    \includegraphics[width=\columnwidth]{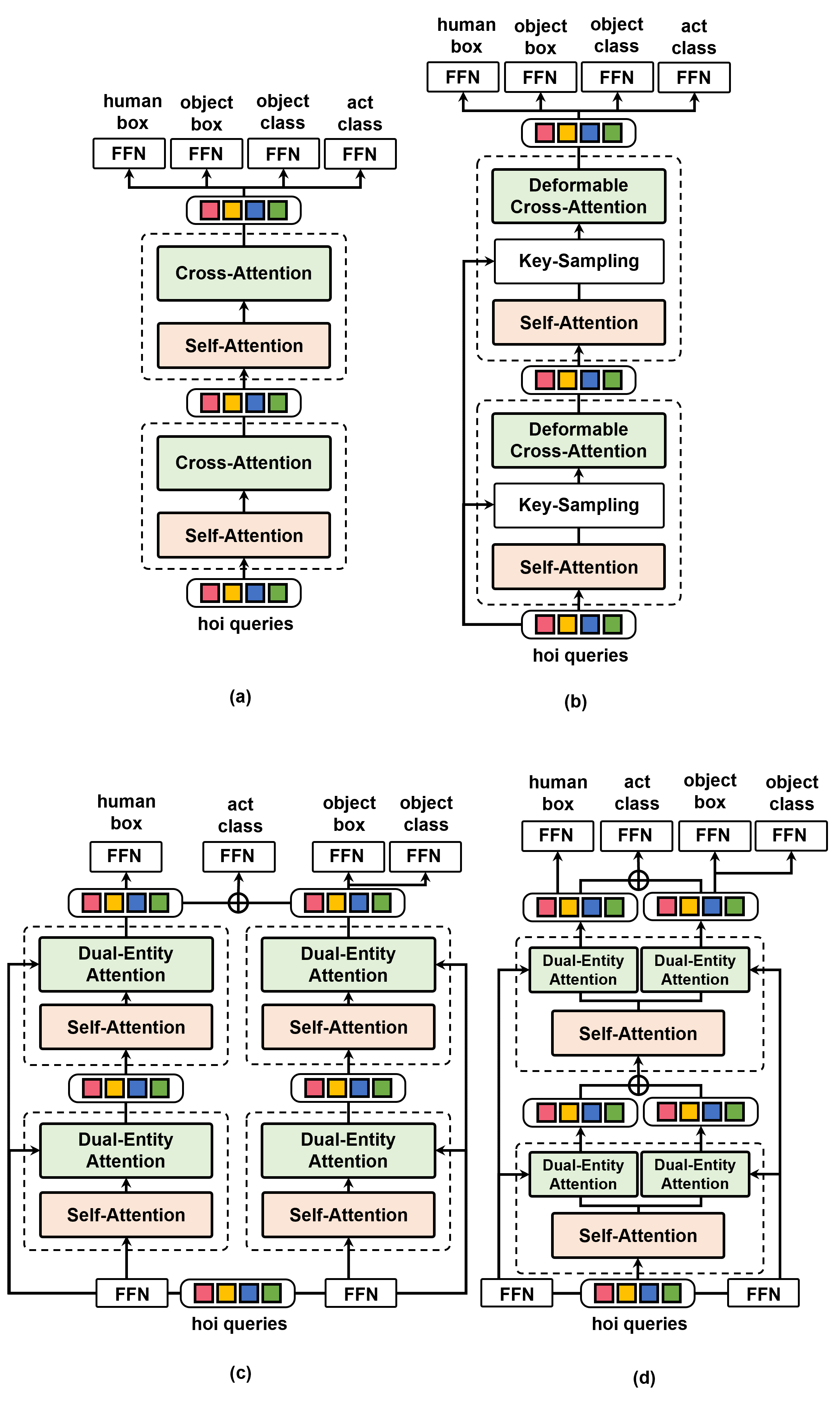}
    \caption{
    Comparison of a simple 2-layer Decoder architecture for: (a) QPIC, and (b) Direct application of Deformable DETR on QPIC, (c) Dual-Entity attention with two streams of decoder layers and (d) Dual-Entity attention that shares the self-attention layer.
    }
    \label{fig:dual}
\end{figure}
In Figure~\ref{fig:dual}, we explore different architectures for Dual-Entity attention.
We start with the most basic form: (a) is the architecture of QPIC, and (b) shows a straightforward application of the deformable attention~\cite{zhu2020deformable} to QPIC.
However, as we discussed in our main paper, (b) degrades the performance a lot from (a),
because unlike its counterpart in object detection, multiple localizations need to be entangled to a single reference point in architecture (b).
Therefore, we first use Dual-Entity attention to disentangle sampling points and attention weights for the participating entities (\ie, human and object), respectively, to improve HOI detection performance.
In Figure~\ref{fig:dual}, (c) and (d) shows two options of dealing with the dual semantics obtained from dual reference points (each for humans and objects).
In (c), each reference point is dealt with a separate stack of decoder layers (\ie, Double-stream), while in (d) they are handled within a single-stream by sharing the self-attention layer where the input is simply the sum of the multiple semantics from the previous decoder layer.
In Table~\ref{tab:merge}, we show that our Dual-Entity attention shows a valid improvement (see (d) vs. (b)), while it even shows better performance than (c) requiring twice the number of decoder parameters.
\begin{table}[h!]
  \centering
  \small
  \begin{tabular}{l | c }
    \toprule
    Method & Default (Full) \\ \hline
    (a) QPIC & 29.07 \\
    (b) QPIC + Deformable attention~\cite{zhu2020deformable} & 27.52 \\
    (c) Double-stream & 28.15 \\
    \textbf{(d) Dual-Entity attention} & \textbf{28.30} \\
    \bottomrule
  \end{tabular}
  \caption{
  Comparison of Dual-Entity attention performance (d) against architecture in Figure~\ref{fig:dual} (a-c). 
  }
  \label{tab:merge}
\end{table}

\begin{figure}
    \centering
    \includegraphics[width=\columnwidth]{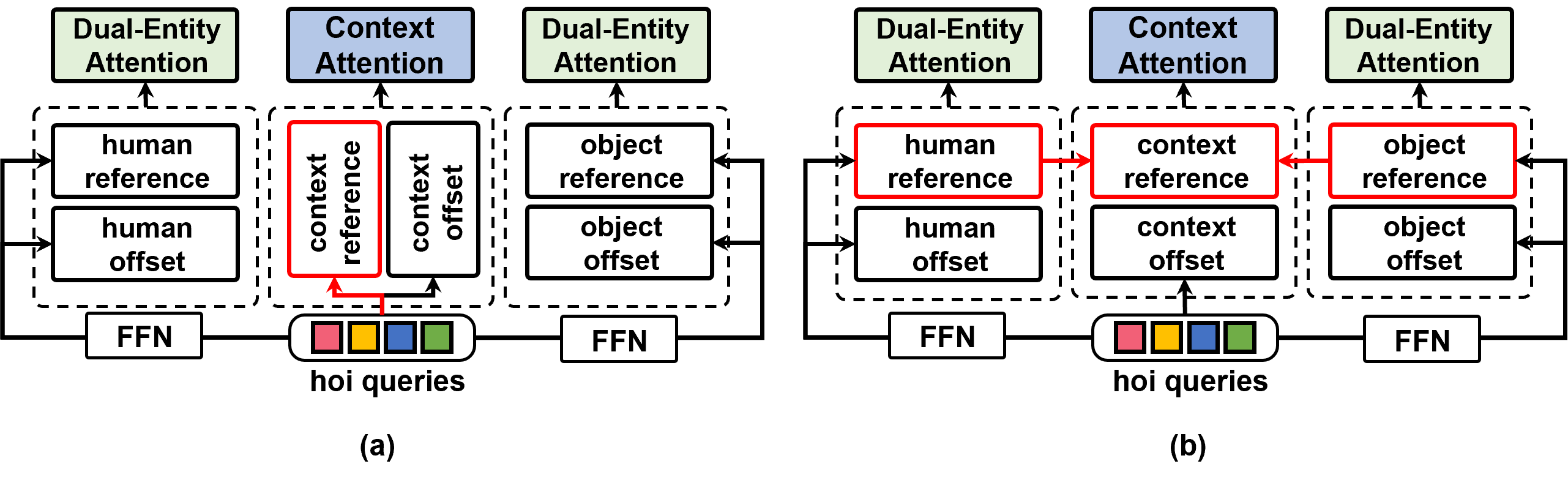}
    \caption{
    Comparison of: (a) context sampling with deformable attention, and (b) Entity-conditioned Context attention.
    }
    \label{fig:ecc}
\end{figure}

\paragraph{Modeling Conditional Context attention.}
In HOI detection, contextual information often gives an important clue in identifying interactions.
In Table~\ref{tab:ecc}, we study the two different methods of obtaining context attention using (a) standard deformable attention and (b) our Entity-conditioned Context attention; note that in standard deformable attention, context reference points are directly obtained from HOI queries with a linear projection while our method conditionally obtain it from human and object reference points (see Figure~\ref{fig:ecc}).
It can be observed that despite its simple structure and minimal delay, our Entity-conditioned Context attention achieves an +0.78p improvement compared to its counterpart.
This implies that the guidance by human and object points is important to effectively model contextual information.
\begin{table}[h!]
  \centering
  \small
  \begin{tabular}{l | c c }
    \toprule
    Method & Default (Full) \\ \hline
    (a) Standard Deformable attention & 29.36 \\
    \textbf{(b) Entity-conditioned Context attention} & \textbf{30.14} \\
    \bottomrule
  \end{tabular}
  \caption{
  Comparison of the performance of Entity-conditioned Context attention against standard deformable attention~\cite{zhu2020deformable}.
  Both (a) and (b) leverage Dual-Entity attention and follow the architectural design of Figure~\ref{fig:merge} (a) for fair comparison.
  }
  \label{tab:ecc}
\end{table}

\begin{figure}
    \centering
    \includegraphics[width=\columnwidth]{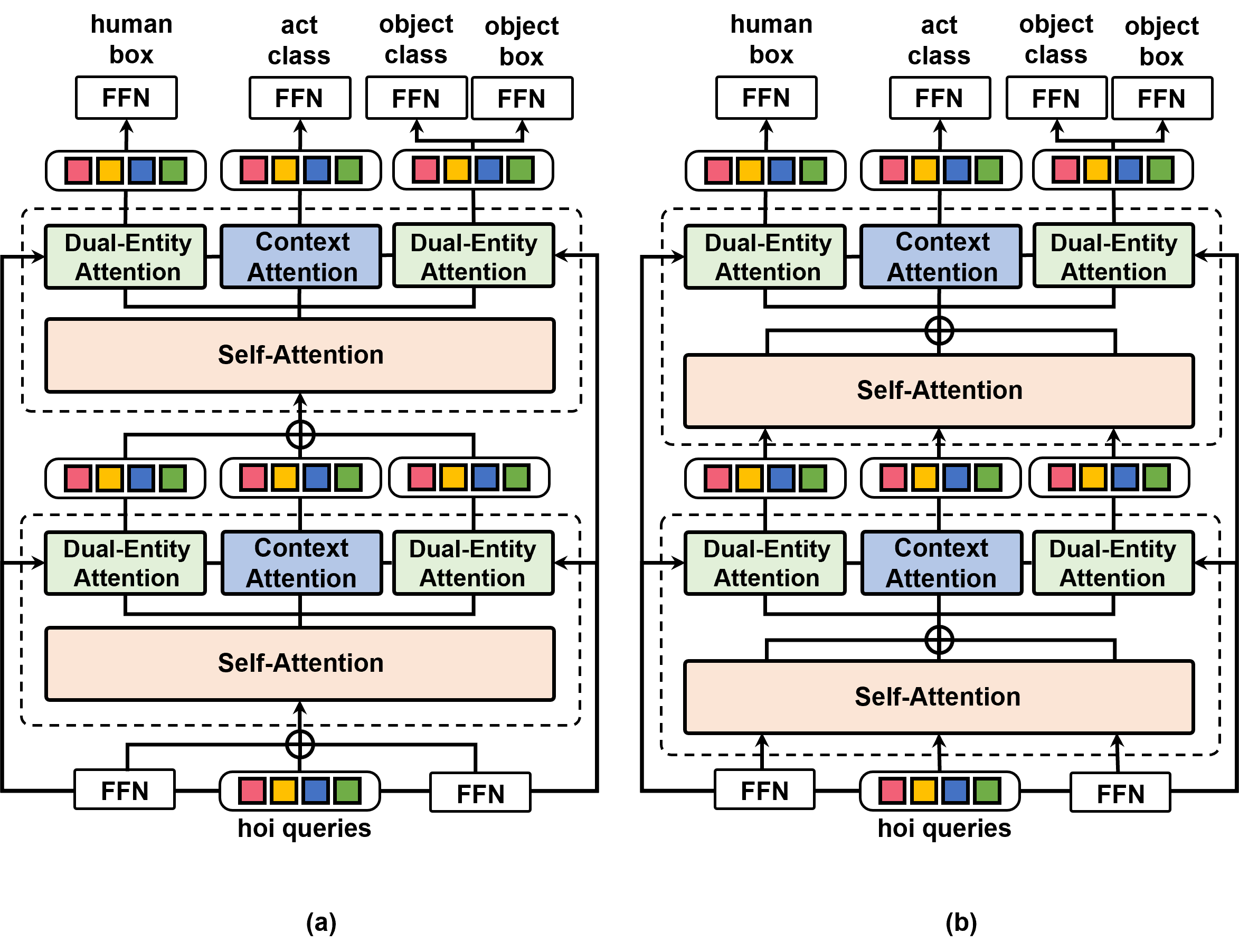}
    \caption{
    Comparison of a simple 2-layer Decoder architecture for Transformer-based HOI detectors: (a) Merging the input of the self-attention, and (b) architecture of MSTR (merging the output of self-attention).
    }
    \label{fig:merge}
\end{figure}

\paragraph{Merging the semantics.}
Figure~\ref{fig:merge} shows two different ways of how to merge the three semantics obtained from our Dual-Entity attention and Entity-conditioned Context attention.
In MSTR, we merge the multiple semantics after applying self-attention separately to each of the semantic features obtained in the previous layer (Figure~\ref{fig:merge} (b)) instead of forcedly composing the input features of the self-attention layer (Figure~\ref{fig:merge} (a)).
Table~\ref{tab:merge} shows that MSTR architecture (b) outperforms (a) by a margin of +1.03p, achieving the final performance.
Note that while (b) is better, MSTR outperforms competing algorithms (presented in Table 2 of main paper) even with architecture (a). 
\begin{table}[h!]
  \centering
  \small
  \begin{tabular}{l | c c }
    \toprule
    Method & Default (Full) \\ \hline
    (a) Merge self-attention input & 30.14 \\
    \textbf{(b) Merge self-attention output} & \textbf{31.17} \\
    \bottomrule
  \end{tabular}
  \caption{
   Comparison of a simple 2-layer Decoder architecture for Transformer-based HOI detectors: (a) Merging the input of the self-attention, and (b) architecture of MSTR (merging the output of self-attention).
  }
  \label{tab:merge}
\end{table}
\subsection{Implementation Details}
Following implementation details in Deformable DETR~\cite{zhu2020deformable}, we use ImageNet pre-trained ResNet-50~\cite{he2016deep} as our backbone CNN and extract multi-scale feature maps without FPN.
The number of attention heads and sampling offsets for deformable attentions are set to $M=8$ and $K=4$, respectively.
The AdamW optimizer is used with the initial learning rate of 2e-4 and weight decay of 1e-4.
All transformer weights are initialized with weights pre-trained in MS-COCO.
For a fair comparison with QPIC~\cite{tamura2021qpic}, we use only 100 HOI queries instead of using 300 ones as in Deformable DETR~\cite{zhu2020deformable}.

\subsection{Details on Datasets and Metrics}
We evaluate our model on two widely-used public benchmarks: the V-COCO (\textit{Verbs in COCO})~\cite{gupta2015visual} and HICO-DET~\cite{chao2018learning} datasets.
V-COCO is a subset of COCO composed of 5,400 trainval images and 4,946 test images.
For V-COCO dataset, we report the $\text{AP}_{\text{role}}$ over $25$ interactions in two scenarios.
In Scenario 1 (denoted as  $\text{AP}_{\text{role}}^{\#1}$), detectors should predict an output indicating the non-existence of an object ([0,0,0,0]) when the target object is occluded, while in Scenario 2 (denoted as $\text{AP}_{\text{role}}^{\#2}$), only the localization of human and interaction classification is scored for such cases.
HICO-DET contains 37,536 and 9,515 images for each training and test splits with annotations for 600 $\langle verb, object \rangle$ interaction types.
In HICO-DET dataset, there are two different evaluation settings: \textit{Default} and \textit{Known object}.
The former measures AP on all the test images, while the latter only considers the images with the object class corresponding to each AP.
We report our score with both settings.
Note that the \textit{Default} is a more challenging setting as we also need to distinguish background images.
We follow the previous settings and report the mAP over three different category sets: (1) all 600 HOI categories in HICO (Full), (2) 138 HOI categories with less than 10 training instances (Rare), and (3) 462 HOI categories with 10 or more training instances (Non-Rare).

\subsection{Training Details of MSTR}
\label{sec:training}
In this section, we explain the details of MSTR training.
MSTR follows a set prediction approach as in previous transformer-based HOI detectors~\cite{kim2021hotr,chen2021reformulating,zou2021end,tamura2021qpic}.
We first introduce the cost matrix of Hungarian Matching for unique matching between the ground-truth HOI triplets and HOI set predictions.

\paragraph{Hungarian Matching for HOI Detection.}
MSTR predicts a fixed number $K$ of HOI triplets that consist of a human box, object box, and binary classification for the $a$ types of actions (where $a$=25 in V-COCO and $117$ for HICO-DET).
Each prediction captures a unique $\langle$human,object$\rangle$ pair with multiple interactions.
$K$ is set to be larger than the typical number of interacting pairs in an image (in our experiment, $K=100$).
Let $\mathcal{Y}$ denote the set of ground truth HOI triplets and $\hat{\mathcal{Y}}=\{\hat{y_i}\}_{i=1}^{K}$ as the set of $K$ predictions.
As $K$ is larger than the number of unique interacting pairs in the image, we consider $\mathcal{Y}$ also as a set of size $K$ padded with $\varnothing$ (there are no ground-truth that matches the prediction). 
Let $y=(b^h,b^o,c^o,a)$ where the ground-truth interaction $y_i$ consists of $b^h_i$ and $b^o_i$ which denotes the normalized coordinates for the interacting human/object box, $c^o_i$ denotes the target object class. and $a_i$ denotes the one-hot for multiple actions.
To find a bipartite matching between these two sets we search for a permutation of $K$ elements $\sigma\in\mathfrak{S}_K$ with the lowest cost:
\begin{equation}
    \hat{\sigma}=\mathop{\mathrm{argmin}}_{\sigma\in\mathfrak{S}_K}\sum_{i}^{K}\mathcal{C}_{\text{match}}(y_i,\hat{y}_{\sigma(i)}),
\label{eq:eq_3}
\end{equation}
where $\mathcal{C}_{\text{match}}$ is a pair-wise \textit{matching cost} between ground truth $y_i$ and a prediction with index $\sigma(i)$.
Now, the ground-truth is written as $y_{i} = (b_i^h,b_i^o,c_i^o,a_i)$ and the prediction is written as $\hat{y}_{\sigma(i)} = (\hat{b}_{\sigma(i)}^h,\hat{b}_{\sigma(i)}^o,\hat{c}_{\sigma(i)}^o,\hat{a}_{\sigma(i)})$ where $\hat{y}_{\sigma(i)}$ is the prediction that has the minimal matching cost with $y_i$.
$\hat{b}_{\sigma(i)}^h$ and $\hat{b}_{\sigma(i)}^o$ are the normalized box coordinates for humans and objects, respectively, $\hat{c}^o_{\sigma(i)}$ is the classification for the target object of the interaction, and $\hat{a}_{\sigma(i)}$ is the predicted actions.

\paragraph{Final Cost/Loss function for MSTR.}
Based on $\mathcal{C}_{\text{match}}$, we calculate the final loss function for all pairs matched.
The cost/loss function for the HOI triplets consists of the localization loss, object classification loss, and the action classification loss as $\mathcal{L}_{\text{H}}=\mathcal{L}_{\text{loc}}+\mathcal{L}_{\text{cls}}+\mathcal{L}_{\text{act}}$ where each function is written as
\begin{align}
\begin{split}
    &\mathcal{L}_{\text{loc}}=\sum_{i=1}^K\big[\mathcal{L}_{\text{loc}}(b^h_i, \hat{b}^h_{\sigma(i)})+\mathcal{L}_{\text{loc}}(b^o_i, \hat{b}^o_{\sigma(i)})\big],\\
    &\mathcal{L}_{\text{cls}}=\sum_{i=1}^K\text{BCELoss}(c_i,\hat{c}_{\sigma(i)}),\\
    &\mathcal{L}_{\text{act}}=\sum_{i=1}^K\text{BCELoss}(a_i,\hat{a}_{\sigma(i)}).
\end{split}
\end{align}
Identical to previous works~\cite{carion2020end,zhu2020deformable,kim2021hotr,chen2021reformulating,zou2021end,tamura2021qpic}, the localization loss is defined by the weighted sum of the L1-loss and the gIoU loss.
\subsection{Convergence speed}
\begin{figure}
    \centering
    \includegraphics[width=\columnwidth]{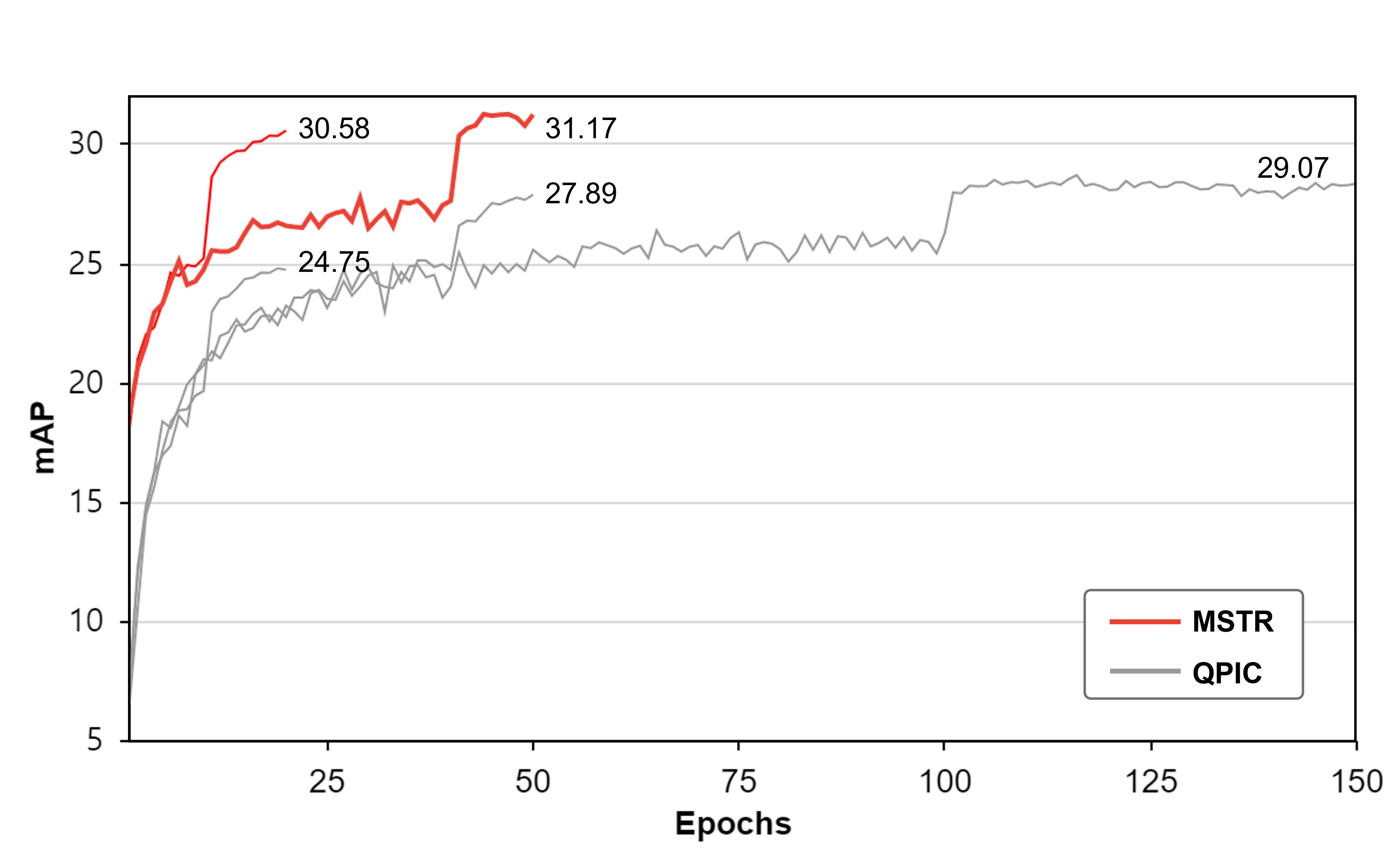}
    \caption{Comparison of convergence curves of QPIC and MSTR in the HICO-DET dataset.
    MSTR shows faster convergence than QPIC under various training schedules for both methods.
    }
    \label{fig:convergence}
\end{figure}

One of the advantages that deformable attention provides is the fast convergence at training.
Figure~\ref{fig:convergence} shows the convergence curve of MSTR compared to QPIC.
Specifically, MSTR requires a much short number of epochs (50 epochs) compared to QPIC (150 epochs) to reach its best score.
Note that MSTR achieves a competitive score to QPIC only with 20 epochs, outperforming QPIC with approximately $\times 4$ shorter training time.
\subsection{Qualitative Analysis for MSTR}
\label{sec:qual}
\begin{figure*}
    \centering
    \includegraphics[width=\textwidth]{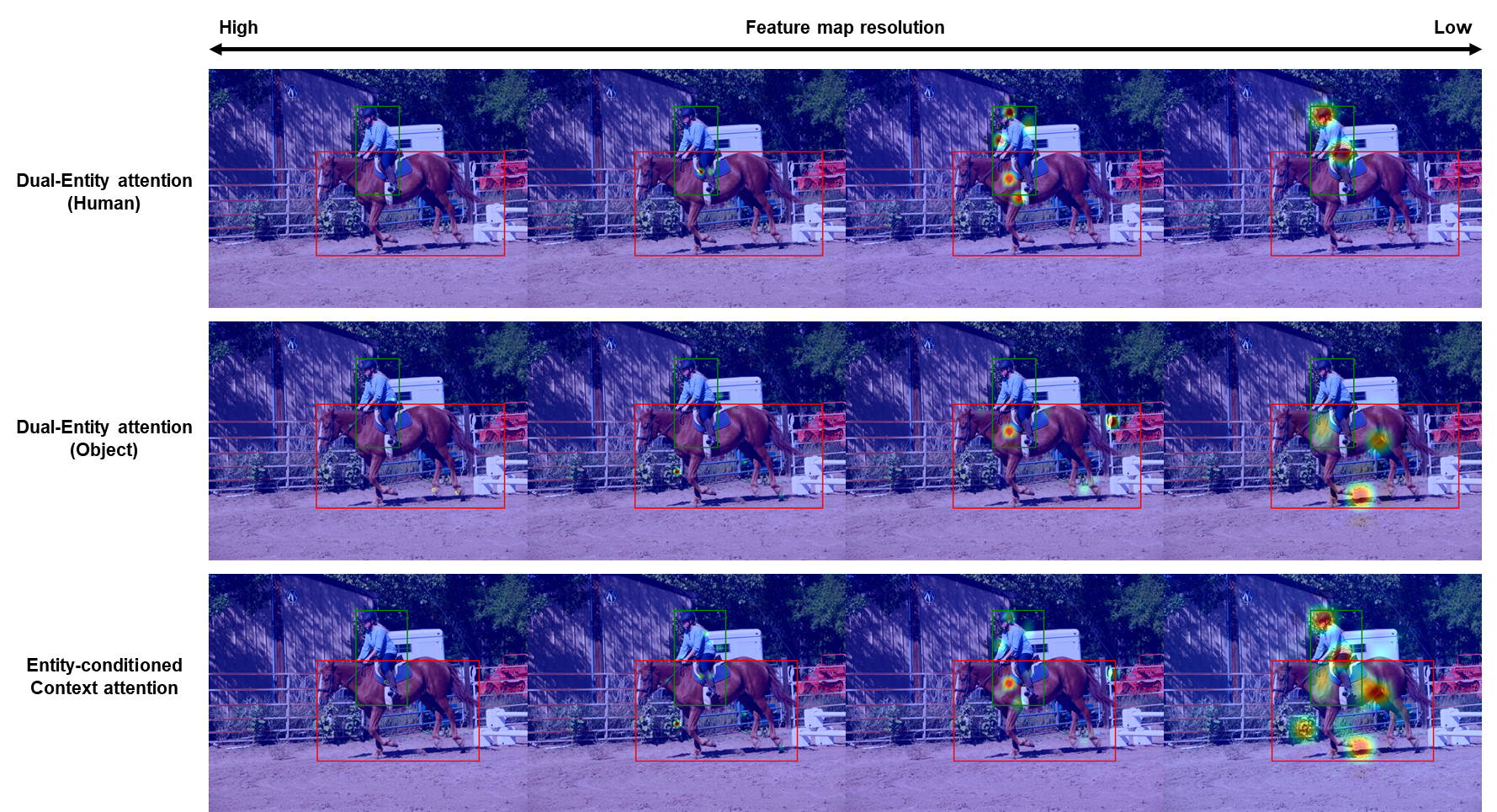}
    \caption{Visualization of the attention for the Dual-Entity attention and Entity-conditioned Context attention of MSTR in multi-scale feature maps for \textit{adjacent} interaction: \textit{ride}.
    }
    \label{fig:qualitative_adjacent}
\end{figure*}

\begin{figure*}
    \centering
    \includegraphics[width=\textwidth]{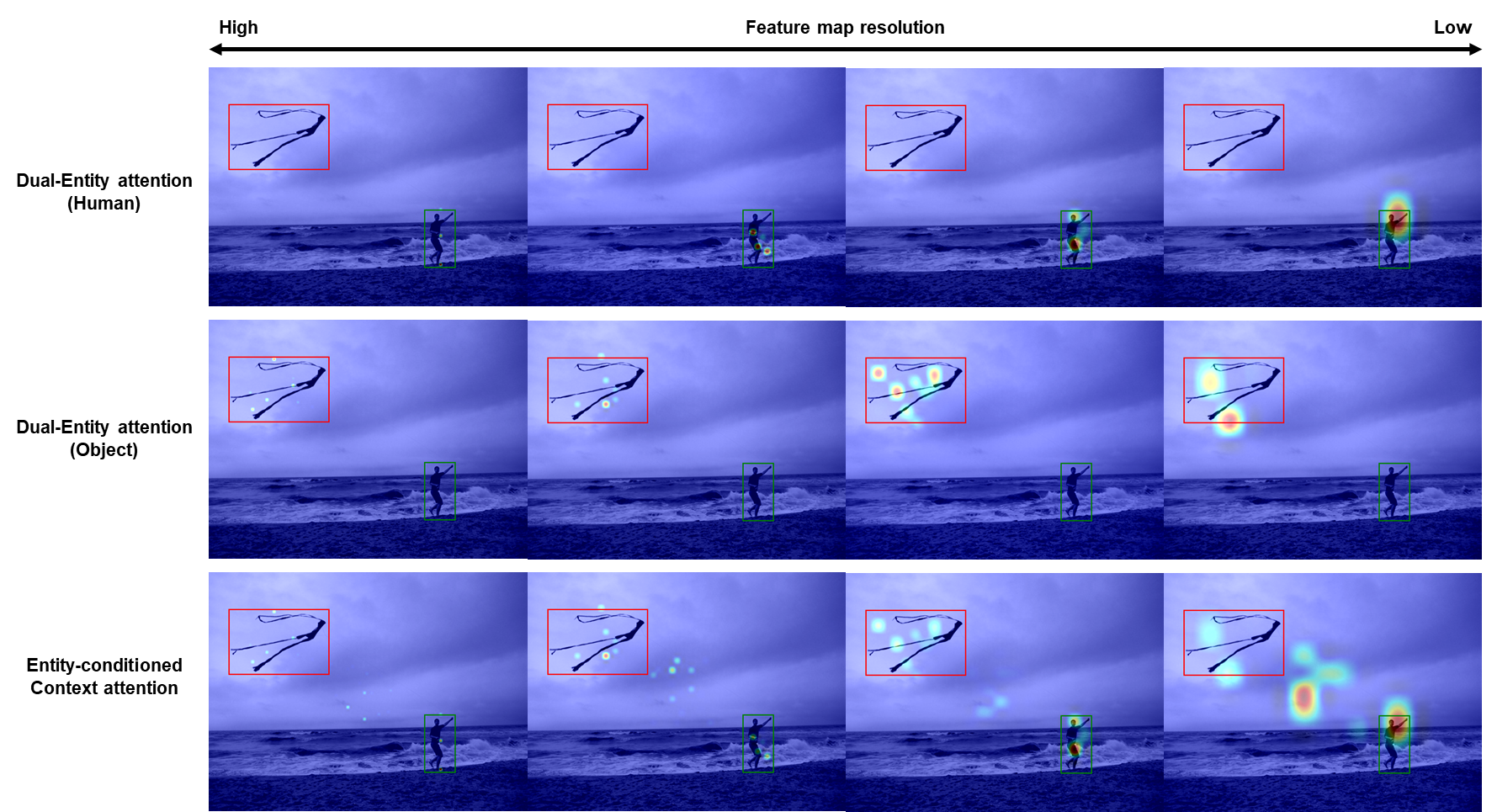}
    \caption{Visualization of the attention for the Dual-Entity attention and Entity-conditioned Context attention of MSTR in multi-scale feature maps for \textit{remote} interaction: \textit{fly}.
    It can be seen that in both \textit{adjacent} interaction and \textit{remote} interaction, MSTR successfully captures the multiple semantics of the human, object, and contextual information across the multi-resolution feature maps.
    }
    \label{fig:qualitative_remote}
\end{figure*}
\begin{figure}
    \centering
    \includegraphics[width=\columnwidth]{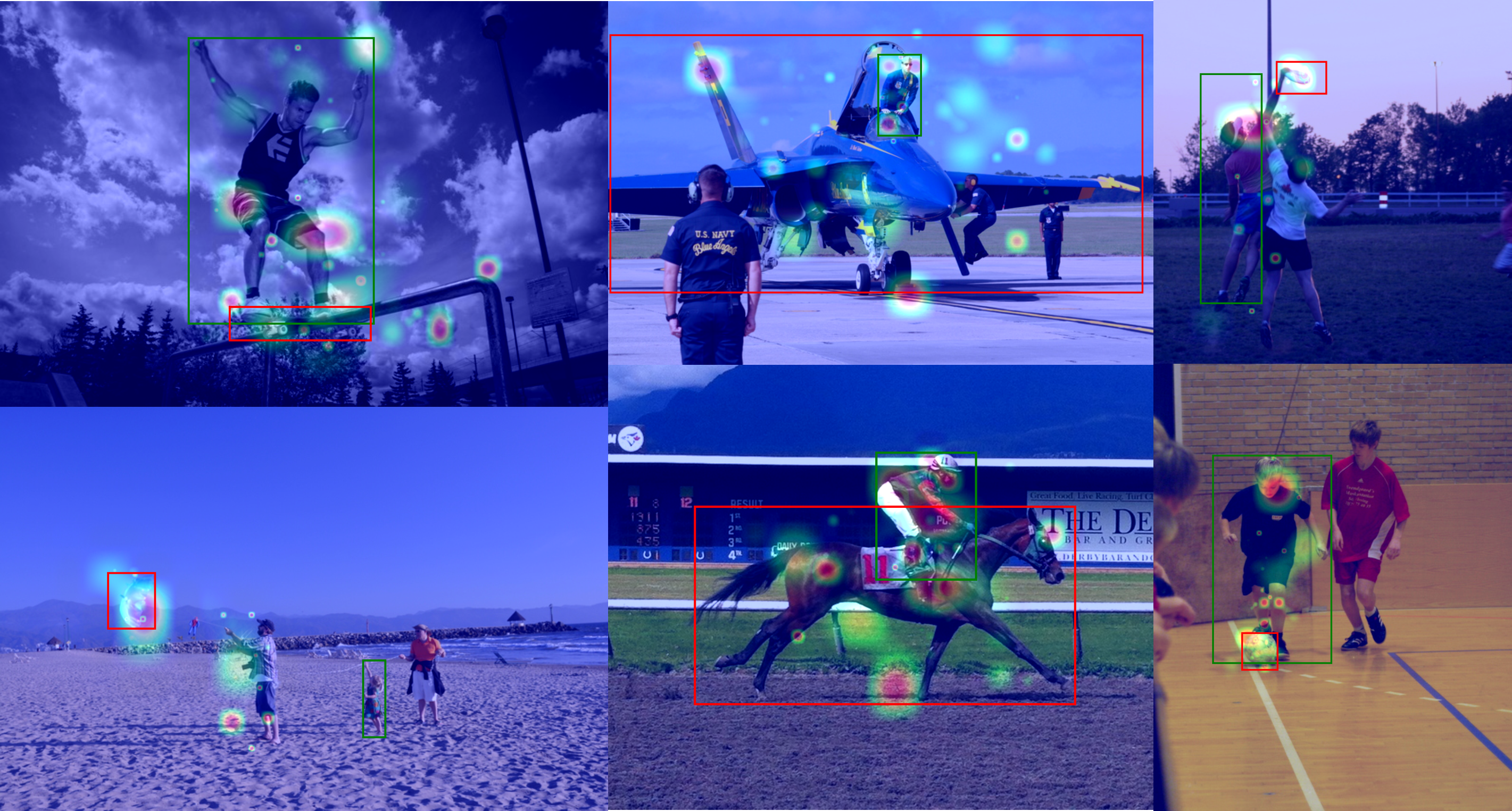}
    \caption{MSTR attentions (Dual-Entity attention and Entity-conditioned Context attention) of different scales all visualized at once.}
    \label{fig:once}
\end{figure}
In this section, we conduct extensive qualitative analysis of MSTR to observe how Dual-Entity attention and the Entity-conditioned Context attention capture different semantics for interactions in a multi-scale environment.

\paragraph{MSTR attentions on multi-scale feature maps.}
We conduct a qualitative analysis of MSTR on both Dual-Entity attention and the Entity-conditioned Context attention in HOI detection to observe how MSTR captures interactions.
Figure~\ref{fig:qualitative_adjacent} shows the visualization of each attention in an \textit{adjacent} interaction: \textit{ride}.
Figure~\ref{fig:qualitative_remote} shows the visualization of each attention in an \textit{remote} interaction: \textit{fly}.
For both cases, we can see that the Dual-Entity attention captures the appearance of the human and object across multiple scales of feature maps. In contrast, the Entity-conditioned Context attention tends to capture an inclusive area that covers both two regions and their intermediate background, effectively capturing the context of the interaction.

\paragraph{MSTR attentions on multi-scale feature maps.}
In Figure~\ref{fig:once}, we provide more qualitative visualizations for the multi-scale attentions of MSTR in various scenes with 1) large human and small object, 2) small human and large object, 3) distant interactions, 4) adjacent interactions.
\subsection{Limitations}
\label{sec:lim}
The main limitation of our work is the bottleneck caused by the extensive size of the \textit{query} element (multi-scale image features, there are about $\times 20$ more image tokens to process compared to the single-scale feature map).
Despite our proposed deformable attentions, MSTR suffers from an estimated $10\%$ increase in parameters and $\sim\times 2$ GFLOPs compared to the single-scale baseline, QPIC~\cite{tamura2021qpic}.
Although recent related works have tackled the efficiency problem in deformable attentions by sampling the query element as well~\cite{wang2021pnp}, the research scope of this work did not cover this issue.

\end{document}